\documentclass[journal,twoside,web]{ieeecolor}
\usepackage{tmi}

\usepackage{cite}
\usepackage{amsmath,amssymb,amsfonts}
\usepackage{algorithmic}
\usepackage{graphicx}
\usepackage{textcomp}
\usepackage{booktabs}

\usepackage[T1]{fontenc}
\usepackage{xcolor}

\usepackage{subfig}

\usepackage{tikz}
\usepackage{pgfplots}
\pgfplotsset{compat=1.18}

\usepackage{hyperref}

\newcommand{\code}[1]{\texttt{#1}}
\newcommand{\invivo}{{\em in vivo}} 
\newcommand{\data}{STIR} 
\newcommand{\exvivo}{{\em ex vivo}}
\newcommand{\Invivo}{{\em In vivo}} 
\newcommand{\Exvivo}{{\em Ex vivo}}  
\def\BibTeX{{\rm B\kern-.05em{\sc i\kern-.025em b}\kern-.08em
    T\kern-.1667em\lower.7ex\hbox{E}\kern-.125emX}}

\begin{document}
\title{Surgical Tattoos in Infrared: A Dataset for Quantifying Tissue Tracking and Mapping} \author{Adam Schmidt, \IEEEmembership{Member, IEEE}, Omid Mohareri, \IEEEmembership{Member, IEEE}, Simon DiMaio, \IEEEmembership{Member, IEEE}, Septimiu E. Salcudean, \IEEEmembership{Fellow, IEEE}
\thanks{This paper was submitted on Nov. 28, 2023.
This work was supported in part by Intuitive Surgical.}
\thanks{Adam Schmidt and Septimiu E. Salcudean are with The Department of Electrical Engineering at The University of British Columbia, Vancouver, BC V5T-2M9 Canada (e-mail: adamschmidt@ece.ubc.ca)}
\thanks{Omid Mohareri and Simon DiMaio are with Advanced Research, Intuitive Surgical, Sunnyvale, CA 94086, USA}}

\maketitle

\begin{abstract}
Quantifying performance of methods for tracking and mapping tissue in endoscopic environments is essential for enabling image guidance and automation of medical interventions and surgery.
Datasets developed so far either use rigid environments, visible markers, or require annotators to label salient points in videos after collection.
These are respectively: not general, visible to algorithms, or costly and error-prone.
	We introduce a novel labeling methodology along with a dataset that uses said methodology, Surgical Tattoos in Infrared (\data{}).
	\data{} has labels that are persistent but invisible to visible spectrum algorithms.
	This is done by labelling tissue points with IR-fluorescent dye, indocyanine green (ICG), and then collecting visible light video clips.
	\data{} comprises hundreds of stereo video clips in both \invivo{} and \exvivo{} scenes with start and end points labelled in the IR spectrum.
	With over 3,000 labelled points, \data{} will help to quantify and enable better analysis of tracking and mapping methods.
	After introducing \data{}, we analyze multiple different frame-based tracking methods on \data{} using both 3D and 2D endpoint error and accuracy metrics.
    \data{} is available at \url{https://dx.doi.org/10.21227/w8g4-g548}
\end{abstract}


\section{Introduction}
\IEEEPARstart{I}{n} surgical robotics, tracking motion of tissue surfaces is important for enabling downstream applications that require understanding of tissue motion and deformation.
These motions can occur due to patient motion such as breathing, or surgical actions, such as retraction or dissection.
A brief list of application in which tissue tracking is important includes: Visual Simultaneous Localization and Mapping (VSLAM) for view expansion~\cite{bergenStitchingSurfaceReconstruction2016}, coverage estimation in colonoscopy screening~\cite{zhangLightingEnhancementAids2021a}, image guidance to maintain registration and tumor locations~\cite{haouchineAccurateTrackingLiver2014}, and automation of tissue scanning~\cite{zhangAutonomousScanningEndomicroscopic2017}.
The importance of having clinically relevant methods for quantification has only increased as more methods for tracking and mapping are implemented and designed for computer vision.
Augmented reality methods have been used for image guidance, but they are missing robust intraoperative deformation tracking~\cite{bernhardtStatusAugmentedReality2017}.
Similarly, tracking is important for endoscopic navigation to improve safety and precision of diagnosis and treatment~\cite{fuFutureEndoscopicNavigation2021}.
This problem is particularly difficult in surgical environments which often have specularity, smoke, blood, and organ movement.
Tissue tracking methods must account for these while also being clinically viable in terms of accuracy and speed.
To assess the clinical viability of these algorithms, we need a means to quantify performance.

Datasets to evaluate tissue tracking thus far either require hand-labeling in software, placing visible markers onto the tissue surface, or are not collected in deformable environments.
This leaves a gap where we could ideally reduce labelling effort, have a marker that is invisible, and collect data over deformable scenarios.

\begin{figure*}[tb]\includegraphics[width=\textwidth]{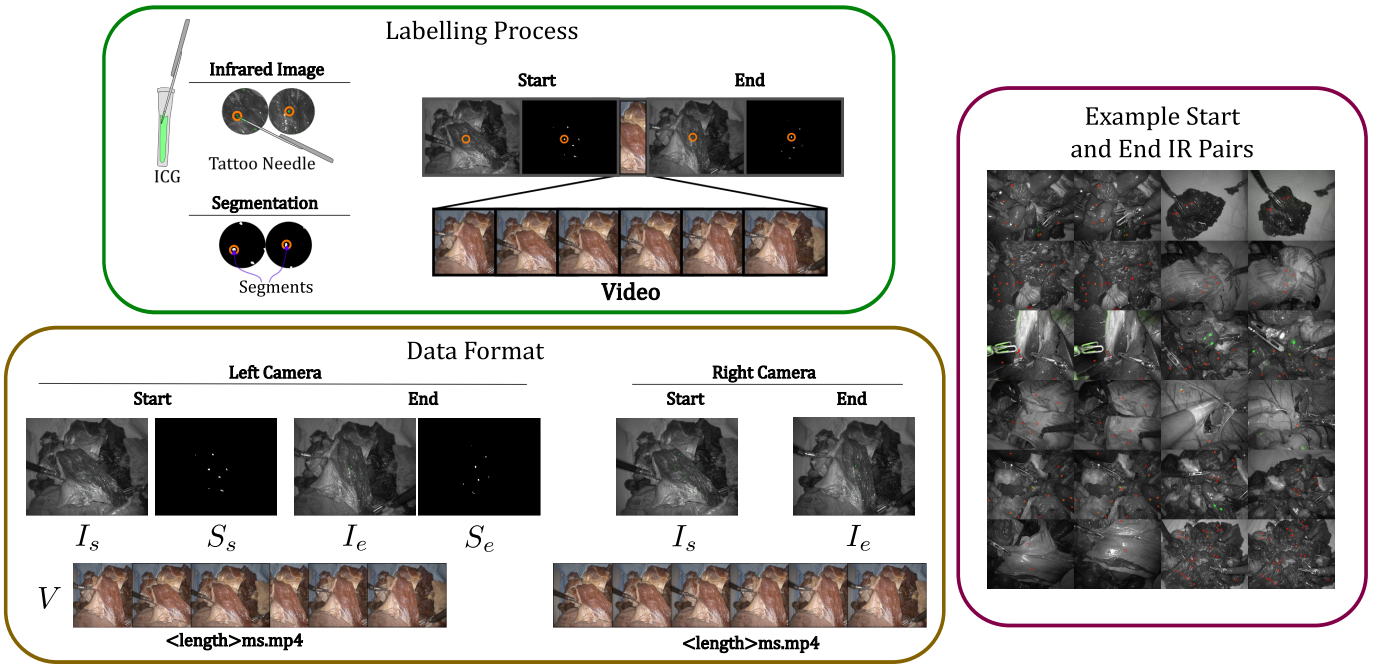}
    \caption{Labelling process: Demonstration of what the labelling process for each clip looks like. Data Format: An example sequence from the released \data{} dataset. Each sequence includes two visible spectrum video clips for the left and right stereo views, IR images at the start and end of the clip, and labels of the start and end segments in the left frame. Specific segments can be seen in the circular cutouts.
	Right: A set of random start and end IR ground truth frame sample pairs from \data{}. The center points of each segment are marked with a red circle.}
    \label{fig:dataformatall}
\end{figure*}

\data{} focuses on doing exactly that; introducing a dataset to help better enable determining accuracy of algorithms.
Our dataset uses IR-labelled points in both \invivo{} and \exvivo{} scenarios. 
\data{} has a over 3000 labelled point motions, and can be used to quantify both 2D and 3D tracking methods.
We note that there are two ways of validating these algorithms and their tracking performance: in 2D image (pixel) space, and in 3D space.
Since \data{} provides rectified stereo data, it can be used for both.
Methods that address monocular tracking or only need to maintain location without a sense of depth (eg. tracking regions for visualization) can use the 2D quantification.
3D is more useful in spaces which need a metric of depth, such as motion compensation~\cite{richaRobust3DVisual2011} or augmented reality~\cite{pelanisEvaluationNovelNavigation2021}.
Refer to Fig~\ref{fig:dataformatall} for a high-level overview of the data labelling process and \data{} as a whole.

We will begin by covering related work in Section~\ref{sec:relatedwork}, providing a more in depth background as we detail what sets \data{} apart.
We then cover experiments in Section~\ref{sec:experiments}, explaining the dataset annotation methodology in Section~\ref{sec:tattooing}.
We cover data collection methods in Section~\ref{sec:datacollection}, and the data processing methodology in Section~\ref{sec:dataprocessing}.
Dataset usage details are provided in Section~\ref{sec:datadetails}.
We perform an evaluation and summarize the results in Section~\ref{sec:evaluationandres}, beginning with a overview of metrics for evaluation in Section~\ref{sec:evaluationmethods}.
In order to provide baselines, we evaluate performance of different frame-to-frame tracking methods (RAFT, CSRT, SENDD) on \data{} in Section~\ref{sec:analysis}.
We conclude with a discussion of what \data{} enables along with limitations and future needs in Section~\ref{sec:discussion}.

In summary, \data{} introduces a novel tissue labelling methodology along with a dataset designed explicitly for the purpose of evaluating algorithm performance in surgical sequences that include tissue deformation.

\section{Related Work}
\label{sec:relatedwork}

In this section we will begin by covering datasets in existing literature.
We first summarize means for indirect quantification, and simulated datasets.
Then we will discuss rigid and nonrigid datasets that can be used to quantify performance in real-tissue environments.
Afterwards we will discuss relevant works which use IR markers.
Finally, we summarize where \data{} stands relative to these works.
We focus on real-tissue data, other resources provide a more broad review of tracking and mapping datasets which include simulated data~\cite{schmidtTrackingMappingMedical2023a}.
For a birds-eye view, Table~\ref{tab:datatable} summarizes the datasets mentioned here that are usable for tissue tracking.

\textbf{Indirect Quantification \& Simulation:}
First, there are indirect means to evaluate tracking by evaluating quality of depth reconstruction.
Some methods evaluate performance indirectly on binocular endoscopic data using stereo depth estimation.
The depth is calculated on the image binocular pairs and Root Mean Squared Error (RMSE) is used as a metric for reconstruction accuracy~\cite{lamarcaDirectSparseDeformable2022,recasensEndoDepthandMotionReconstructionTracking2021}.
These are based on the assumption that the depth estimation used for ground truth is accurate for quantifying deformation.
Since these methods are indirect, we do not discuss them in this paper.
Although simulated training and evaluation data is important, with \data{} we focus on real-tissue scenarios.
Accurate simulation of tissue motion with realistic rendering remains an open research problem~\cite{schuleModelbasedSimultaneousLocalization2022} and cannot be used on its own for algorithm evaluation since algorithms still need to be translated to and evaluated on real environments.
Datasets need to include video sequences to evaluate tracking, so those with standalone frames such as SERV-CT~\cite{edwardsSERVCTDisparityDataset2022} are also of limited relevance here.

\textbf{Rigid datasets:}
Some datasets collect ground truth via collecting depth measurements with structured light in scenes that undergo rigid tissue motion.
These generate a rigid ground truth that can be used for evaluating tissue tracking.
EndoSLAM~\cite{ozyorukEndoSLAMDatasetUnsupervised2021} presents multiple \exvivo{} tissue scenes with thousands of scanned frames.
In the SCARED~\cite{allanStereoCorrespondenceReconstruction2021} dataset, depth is calculated using structured light in a porcine cadaver for a set of keyframes.
This allows for estimating flow on rigid data by using the known depth along with the pose transformations.
In C3VD~\cite{bobrowColonoscopy3DVideo2022}, thousands of depth frames were created and aligned with images collected from videos of a 3D printed phantom.
The primary issue with using rigid models to quantify tissue tracking methods is that depth does not necessarily quantify underlying motion.
Even though better tissue tracking likely correlates with higher depth accuracy, there is not a direct performance relationship between the two.
For example a checkered tablecloth sliding over a table maintains the same depth, but is shifting in motion.
Thus, endoscopic methods need a means to quantify motion in addition to depth.

\begin{table}[tb]
	\caption{Dataset comparison of datasets intended or usable for tissue tracking performance estimation. Some are intended for auxiliary problems such as depth estimation, but can still be used to quantify rigid tissue motion. Datasets below the horizontal line are rigid and can use depth/3D position to evaluate rigid flow. SW: software annotated, Def.: deformable, *: porcine cadaver.}
\begin{center}
\begin{tabular}{l|lllll}
	\toprule
	Dataset & Labelled & Type & Def. & Marker & Point\\
	& Frames &  &  & & Tracks\\
	\cmidrule{1-6}
	\data{} & 576 & {\em in, ex vivo} & Y & IR & 3604\\
	SuPer~\cite{liSuperSurgicalPerception2020} & 52 & \exvivo{} & Y & SW & 20\\ 
	Sem. SuPer~\cite{linSemanticSuPerSemanticawareSurgical2023} & 600 & \exvivo{} & Y & Beads & 240\\
SurgT~\cite{cartuchoSurgTChallengeBenchmark2023} & 24,548 & \invivo{} & Y & SW & 32\\
	\cmidrule{1-6}
	EndoSLAM~\cite{ozyorukEndoSLAMDatasetUnsupervised2021} & 42,700 & \exvivo{} & N & Depth & N/A\\
	SCARED~\cite{allanStereoCorrespondenceReconstruction2021} & 40 & \invivo{}* & N & Depth & N/A \\
	C3VD~\cite{bobrowColonoscopy3DVideo2022} & 10,015 & Phantom & N & Depth & N/A\\
\bottomrule
\end{tabular}
\end{center}
\label{tab:datatable}
\end{table}

\textbf{Nonrigid datasets:}
One way to evaluate tracking in a nonrigid environment is by marking the tissue surface with stitches or beads
that are visible to tracking algorithms.
This makes the scene visually different from the unmarked scenes in surgeries.
Yip et al.~\cite{yipTissueTrackingRegistration2012} quantify motion by using 2mm steel beads to act as markers on the tissue surface.
In Semantic SuPer~\cite{linSemanticSuPerSemanticawareSurgical2023}, green pins are used on the tissue surface for ground truth.

Another means of providing ground truth is to hand annotate points using software, as done in the SurgT challenge~\cite{cartuchoSurgTChallengeBenchmark2023} and SuPer~\cite{liSuperSurgicalPerception2020}.
Labellers select points that are visibly salient and track them over time as they watch a video frame by frame.
These points can be used as ground truth for algorithm evaluation.
However, only points that are salient to the labellers can be used, which could introduce bias.
Also, annotation is difficult, which can make it expensive to collect larger datasets.
That said, there is work that has looked to address this with crowsourcing data collection~\cite{maier-heinCrowdtruthValidationNew2015}.
With all relevant datasets mentioned, Table~\ref{tab:datatable} summarizes the datasets that are usable for tissue tracking.
Finally, we note that there are non-medical datasets such as SINTEL (simulated)~\cite{butlerNaturalisticOpenSource2012} along with the recent PointOdyssey~\cite{zhengPointOdysseyLargeScaleSynthetic2023} and TAP-Vid (hand-annotated)~\cite{doerschTAPVidBenchmarkTracking2022}.

\textbf{Indocyanine Green (ICG) markers:}
In order to robustly evaluate tracking models, we would like to understand motion in ill-featured and textureless regions.
For \data{}, we are inspired by the medical tattooing of large regions that happens in colonoscopy.
For these procedures, India ink or equivalents (SPOT EX tattoo) are used for marking regions to return to in a colonoscopy for later resection.
These markers typically use a large amount of ink (0.5 to 1 ml) injected into tissue with a syringe, and are not designed for accuracy.
Furthermore, these markers are visible and could thus confound a tracking model.
Instead, we want a way to annotate without having the labels be visible under visible light.

Infrared has been used in prior literature for guidance of surgical robots by attaching ICG beads to tissue with cyanoacrylate
~\cite{shademanFeasibilityNearinfraredMarkers2013, geLandmarkGuidedDeformableImage2019} for robots performing anastomosis.
The markers can also be sub-surface with fewer beads for longitudinal studies~\cite{geNovelIndocyanineGreenbased2021}.
These methods are designed for guidance rather than algorithm quantification and thus they do not worry about the markers being visible in the visible light spectrum.
We are instead focused on quantifying tracking methods without marker visibility.

\textbf{STIR:}
Now with the relevant work summarized, we explain \data{} in the context of these.
We create a dataset, \data{}, to evaluate tissue tracking performance.
\data{} is the first work to create a tracking dataset with surgically tattooed tissue with Infrared (IR) ICG markers.
Unlike prior work, this novel methodology is unobtrusive, and can be used to robustly annotate any points on the tissue surface.
The method is convenient and does not require a large labelling effort.
\data{} is large and has both \invivo{} and \exvivo{} scenes collected from a da Vinci Xi surgical robot.
Some resultant tattoos, and the process are shown in Fig.~\ref{fig:dataformatall}.
After tattooing, the data collection process entails capturing a start IR frame, recording a visible light video as we perform actions, and capturing an end IR frame.
These are performed for a large set of actions and scenes.
Then, tracking performance can be evaluated by testing algorithms on the visible clip and evaluating accuracy metrics on the IR segmented regions.
\data{} is significant as it will help evaluate tissue tracking methods, test performance of SLAM methods, and enable reconstruction methods to be quantified using more detailed information than just photometric errors~\cite{schmidtRecurrentImplicitNeural2022,wangNeuralRenderingStereo2022}.
This is important for enabling clinical usage of algorithms.

\begin{figure}[tb]\centering
    \includegraphics[width=0.95\linewidth]{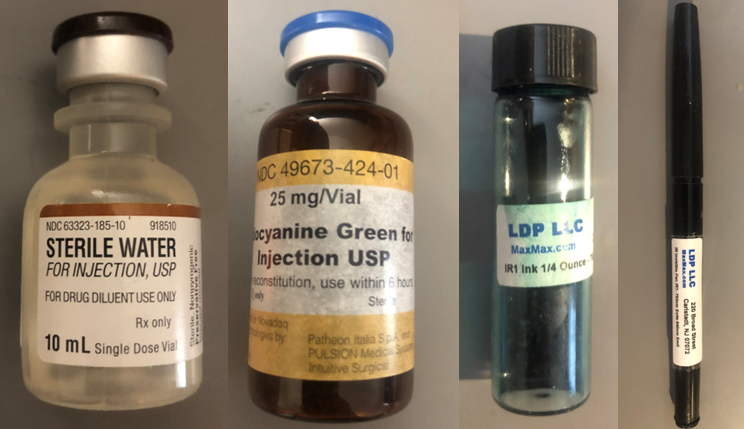}
    \caption{Left to right: sterile water, ICG, IR ink, and IR pen used for labelling tissue.}\label{fig:dyes}\end{figure}

\section{Experiments}
\label{sec:experiments}
\subsection{Surgical Tattooing Methodology}
\label{sec:tattooing}

In order to finally settle on tattooing using a tattoo needle, we experimented with multiple different methods to apply IR ink to surgical tissue.
Initial experiments used IR pens and ink from \textbf{maxmax.com} (see Fig.~\ref{fig:dyes}).
These worked for visibility, but the lack of a dry tissue surface either prevented the marker from releasing ink, or caused the ink to bleed.
We also experimented with IR beads embedded under the tissue surface.
The beads were embedded by cutting a hole into {\em ex vivo} tissue using a scalpel and then sliding the bead in under said hole.
This would be too invasive for {\em in vivo} experiments.
The insertion process has some drawbacks: if the bead is too close to the surface it leaves a bump, or if the bead is too deep it is very diffuse in color.
Another issue is that small scalpel marks left from where the beads are inserted provide additional features that can artificially improve algorithm performance.

\begin{figure}[tb]\centering
    \includegraphics[width=0.95\linewidth]{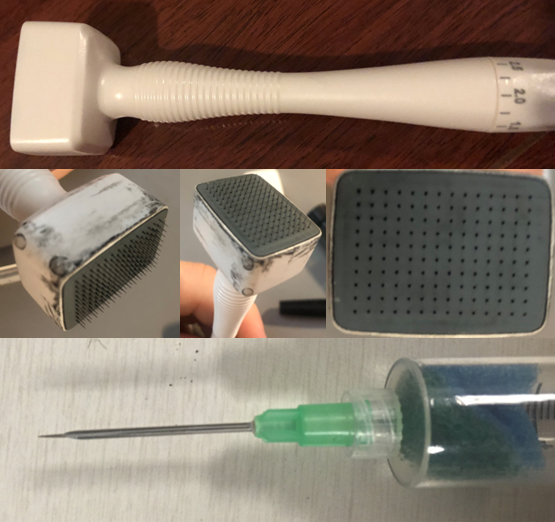}
    \caption{Different means to apply ink to tissue surface. Top: Microneedling device. Middle: Close-ups of needle array on microneedle grid. Bottom: Hypodermic needle with a tattoo needle inserted into the hollow.}\label{fig:applicators}\end{figure}

\begin{figure}[tb]\centering
    \includegraphics[width=0.95\linewidth]{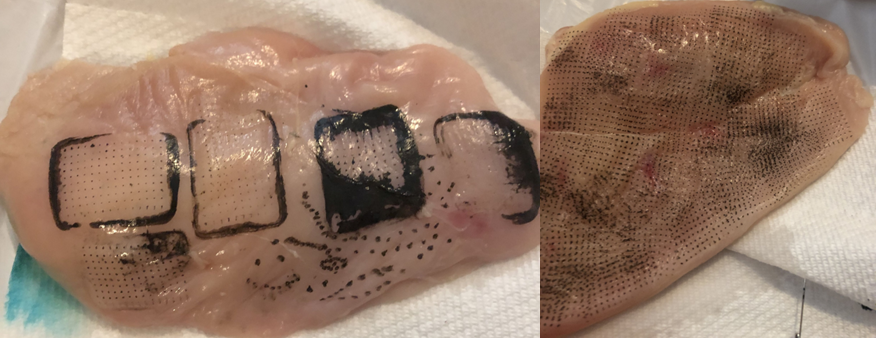}
    \caption{India ink microneedle array tattooing results on a chicken breast. On the left are failed dye applications due to uneven ink. On the right are successful dye applications.}\label{fig:microneedle}\end{figure}

Thus we looked for ways to deposit ink on the tissue surface with needles, microblades, or microneedles.
All of these entail dipping a needle/blade into an ink reservoir, and then lightly applying the ink on the tissue surface.
The microblade (a small linear array of microneedles) was too dense and only left large line-like marks.
The microneedle array tattoo was very clear when the ink was properly applied, but it was difficult to prevent ink from pooling on the plastic body of the device (Fig.~\ref{fig:applicators}).
This made consistently even application of ink into a hard to repeat process.
If these issues can be amended, a custom device could be made which would then provide a promising way to generate much denser annotations.
See Fig. \ref{fig:microneedle} for an example of microneedling success and failure.
For our dataset, \data{}, the microneedle tool and ink application would not fit {\em in vivo} through a cannula (1 to 3cm port in minimally invasive surgery) without substantial changes so we moved to using simple tattoo needles.

\begin{figure}[tb]\centering
    \includegraphics[width=0.95\linewidth]{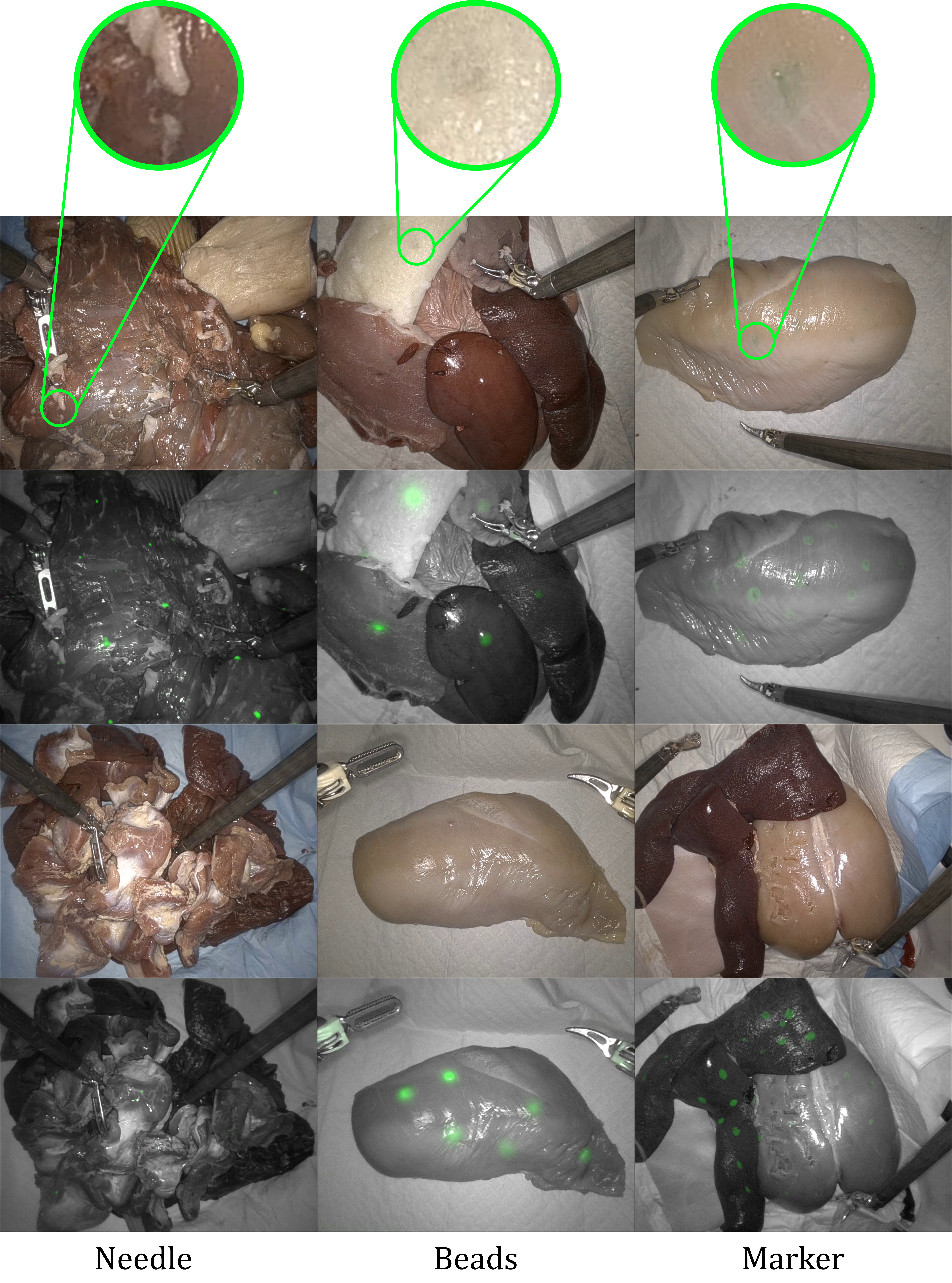}
    \caption{Types of labelling. The tattoo needle, IR-ink marker, and bead insertion method are shown. Two captures of visible/IR pairs are shown for each. Note how the bump of the beads or the ink the marker pen are both visible in the visible spectrum images.}\label{fig:types}\end{figure}

\begin{figure}[tb]\centering
    \includegraphics[width=0.95\linewidth]{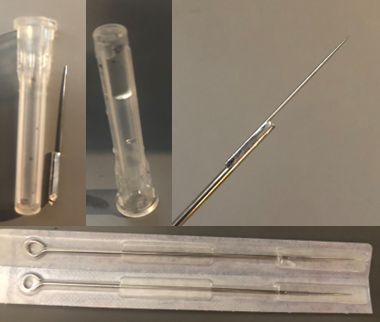}
    \caption{Top: Reservoir with clipped needle, reservoir holding liquid with surface tension, tip of 1RL tattoo needle. Bottom: Packaged tattoo needles.}\label{fig:needle}\end{figure}

After experimenting with different types and sizes of tattoo needles (3RL (3-needle, round), 9RL, etc), we found that the ink surface tension causes the tattooed points to be too large with all but a 1RL (single point) tattoo needle.
Commercially, powered tattoo needles are often used, but this requires electricity, sanitization, and would not fit through the assist cannula for \invivo{} applications.
To ease reapplication of ink, we attempted nesting the tattoo (non-hypodermic, ink coats outside) needle inside a hypodermic needle to allow the ink to be dispensed or to drip onto the tattoo needle (Fig. \ref{fig:applicators}).
Adjusting the plunger proved too difficult, but an automated way to re-coat the tattoo needle would be promising (similar to how powered tattoo needles do with an ink well).
With this evaluation, we decided to use the simplest method of using a separate ink reservoir along with a tattoo needle that can be dipped in the reservoir each time before we tattoo a point.
The reservoir was chosen to be small enough to hold the ink in with surface tension, preventing spills.
This method is small profile and feasible for both \invivo{} and \exvivo{} experiments.
To make the ICG (Indocyanine Green) mixture, we mix 25mg of ICG with 30ml of sterile water.
This is then injected to fill the ink reservoir for tattooing.
Fig.~\ref{fig:types} and
Fig.~\ref{fig:needle} illustrate the reservoir and how each labelling type looks on tissue, with the needle giving the most fine-grained tattooed label points for measurement.

\subsection{Data Collection Experiments}
\label{sec:datacollection}

To create a robust dataset for estimating performance of short-term tracking methods, we include many different types of motion and tissue.
All experiments are performed on the da Vinci Xi robotic surgical system.
\data{} was collected on an IACUC-approved study in an AALAC-accredited facility.  
The da Vinci Xi provides a mode called Firefly that changes the endoscope mode to capture light in the IR spectrum and allows visualization of IR-fluorescent inks such as ICG.
We will explain Firefly mode, as it is essential to our dataset.
Firefly mode enables capture of images using the exact same calibration and camera parameters without any change in pose.
The Firefly system records at 25 fps just as the normal endoscope does and takes 6 frames (\(\sim 0.23\)s) to transition to and from the mode.
ICG is excited at NIR 789nm and emits at 814nm which is shown in Firefly mode but invisible in the visible light mode.
Collecting each tissue motion sequence entails first tattooing the tissue with a large set of points.
Then, after tattooing, an IR start frame is captured by switching to Firefly mode.
A visible light sequence can then be captured by switching modes to white light.
Finally, an IR finish frame is captured by switching to Firefly mode.
This is all recorded in a single video, and separated after the fact using system transition data.

\begin{figure}[tb]\centering
    \subfloat{{\includegraphics[height=5.2cm]{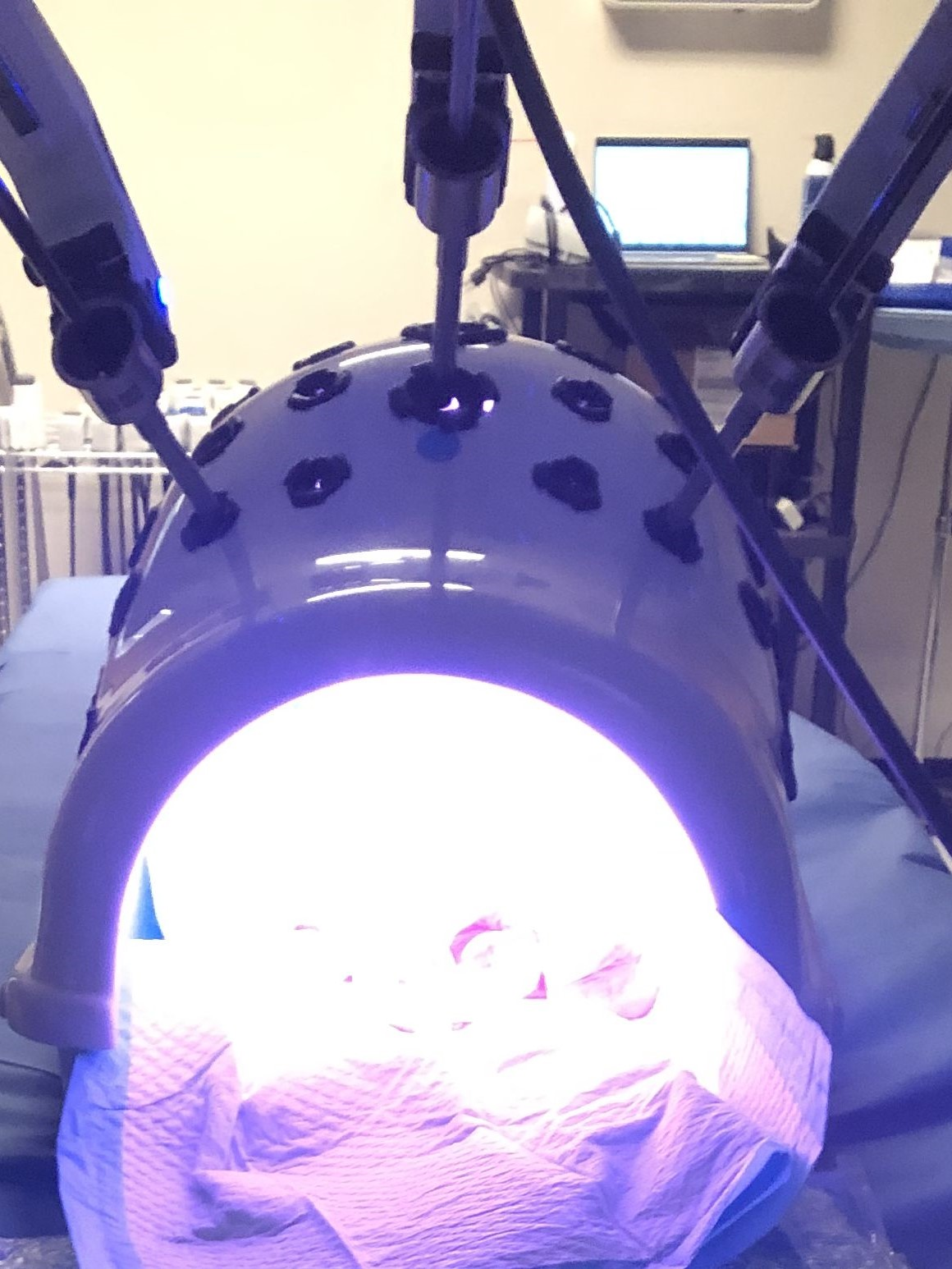} }}\subfloat{{\includegraphics[height=5.2cm]{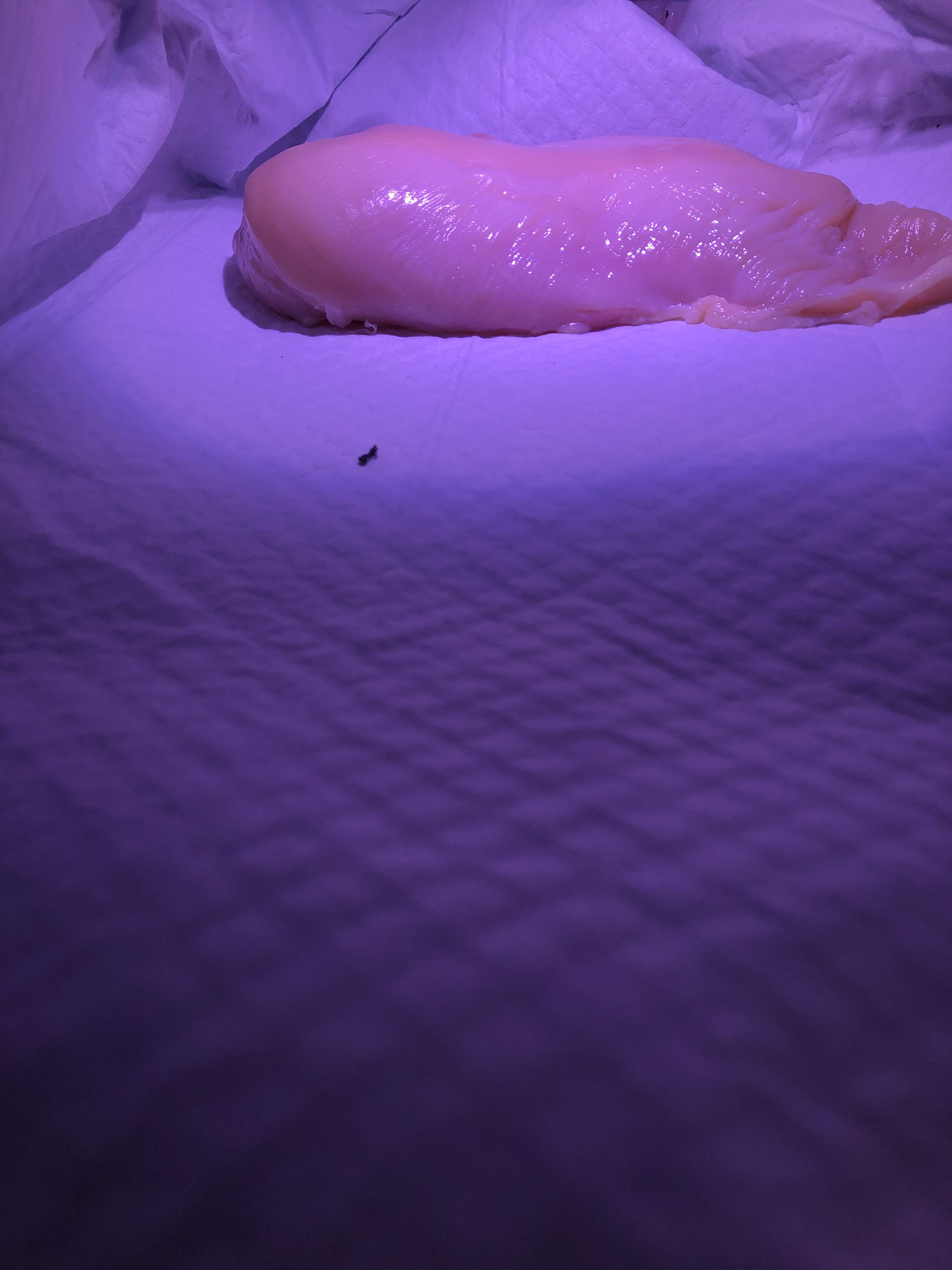} }}\\\caption{Left: Robot docked in the body model under IR illumination. Right: Photo from inside of the body model.}
    \label{fig:bodymodel}
\end{figure}

\begin{figure}[tb]\centering
    \begin{tikzpicture}

\definecolor{chocolate2267451}{RGB}{226,74,51}
\definecolor{dimgray85}{RGB}{85,85,85}
\definecolor{gainsboro229}{RGB}{229,229,229}
\definecolor{lightgray204}{RGB}{204,204,204}

\begin{axis}[
axis background/.style={fill=gainsboro229},
axis line style={black},
height=3.5cm,
tick align=outside,
tick pos=left,
	title={Histogram of time for {\em ex vivo}},
width=0.9\linewidth,
x grid style={lightgray204},
xlabel=\textcolor{dimgray85}{Time in seconds},
xmajorgrids,
xminorgrids,
xmin=0, xmax=63,
xtick style={color=dimgray85},
y grid style={lightgray204},
ylabel=\textcolor{dimgray85}{Number of clips},
ymajorgrids,
yminorgrids,
ymin=0, ymax=431.55,
ytick style={color=dimgray85}
]
\draw[draw=black,fill=chocolate2267451,very thin] (axis cs:0,0) rectangle (axis cs:3,309);
\draw[draw=black,fill=chocolate2267451,very thin] (axis cs:3,0) rectangle (axis cs:6,411);
\draw[draw=black,fill=chocolate2267451,very thin] (axis cs:6,0) rectangle (axis cs:9,190);
\draw[draw=black,fill=chocolate2267451,very thin] (axis cs:9,0) rectangle (axis cs:12,90);
\draw[draw=black,fill=chocolate2267451,very thin] (axis cs:12,0) rectangle (axis cs:15,50);
\draw[draw=black,fill=chocolate2267451,very thin] (axis cs:15,0) rectangle (axis cs:18,25);
\draw[draw=black,fill=chocolate2267451,very thin] (axis cs:18,0) rectangle (axis cs:21,12);
\draw[draw=black,fill=chocolate2267451,very thin] (axis cs:21,0) rectangle (axis cs:24,10);
\draw[draw=black,fill=chocolate2267451,very thin] (axis cs:24,0) rectangle (axis cs:27,6);
\draw[draw=black,fill=chocolate2267451,very thin] (axis cs:27,0) rectangle (axis cs:30,4);
\draw[draw=black,fill=chocolate2267451,very thin] (axis cs:30,0) rectangle (axis cs:33,2);
\draw[draw=black,fill=chocolate2267451,very thin] (axis cs:33,0) rectangle (axis cs:36,4);
\draw[draw=black,fill=chocolate2267451,very thin] (axis cs:36,0) rectangle (axis cs:39,7);
\draw[draw=black,fill=chocolate2267451,very thin] (axis cs:39,0) rectangle (axis cs:42,0);
\draw[draw=black,fill=chocolate2267451,very thin] (axis cs:42,0) rectangle (axis cs:45,3);
\draw[draw=black,fill=chocolate2267451,very thin] (axis cs:45,0) rectangle (axis cs:48,1);
\draw[draw=black,fill=chocolate2267451,very thin] (axis cs:48,0) rectangle (axis cs:51,2);
\draw[draw=black,fill=chocolate2267451,very thin] (axis cs:51,0) rectangle (axis cs:54,3);
\draw[draw=black,fill=chocolate2267451,very thin] (axis cs:54,0) rectangle (axis cs:57,1);
\draw[draw=black,fill=chocolate2267451,very thin] (axis cs:57,0) rectangle (axis cs:60,0);
\end{axis}

\end{tikzpicture}
 \caption{\Exvivo{} experiment clip length distribution}
    \label{fig:histogramexvivo}
\end{figure}
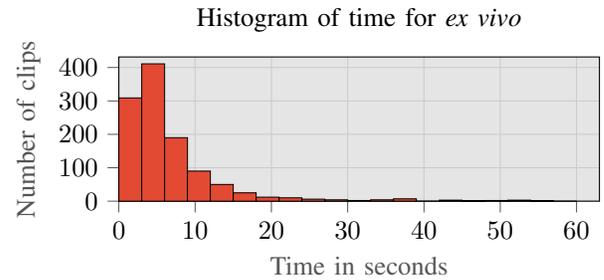

\textbf{\Exvivo{}:}
The factors we considered important for the \exvivo{} data collection comprised collecting a breadth of tissue size, texture, stiffness, color, and features.
Additionally, we extend the tissue types used by placing importance on having multiple tissues samples present to be able to evaluate discontinuity and occlusion.
We do not focus on physically realistic scenes here since we expect the \invivo{} dataset to fit this criterion more.
Instead, we collect a diverse set of samples that can prove difficult to tracking algorithms.
For our \exvivo{} experiments, we select a multitude of tissue types to test everything from well-featured (pork chop) to ill-featured surfaces (liver/heart).
Prior tissue tracking datasets have used~\exvivo{} chicken and steak as tissue before~\cite{liSuperSurgicalPerception2020,linSemanticSuPerSemanticawareSurgical2023}, and we include these tissues while also extending our dataset to include additional tissue types.
Our dataset includes the following tissue types: pork chop, aorta, beef tongue, chicken hearts, pig heart, chicken breast, pork stomach, pork intestine, pork spleen.
To enable a similar lighting environment to that in surgery, we place tissue into a body model.
This is shown in Fig.~\ref{fig:bodymodel}.
After placement of tissue, we begin recording.
The types of tissue actions we consider are the following: bulk movement, stretching and squishing, camera movement, palpation, instrument-tissue and tissue-tissue occlusion, and cutting and tearing.
Collecting a diverse set of actions will help to evaluate performance of algorithms that account for discontinuity and changes.
Although we collect a wide range of actions via maintaining a checklist of these actions for each time we performed collection, we do not prescribe a specific distribution to necessary actions, and let the operator determine which actions to perform for each scene.
Post-hoc labelling of actions could be useful for future work.
Actions are generally intended to be kept below 10 seconds. A histogram of action lengths is shown in Fig.~\ref{fig:histogramexvivo}.

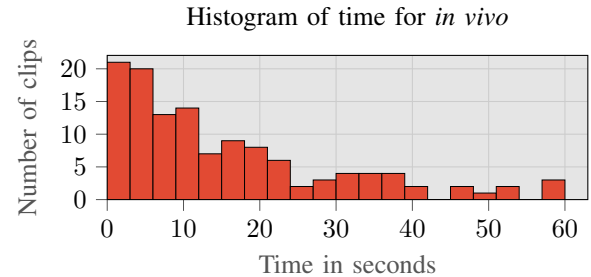
\begin{figure}[tb]\centering
    \begin{tikzpicture}

\definecolor{chocolate2267451}{RGB}{226,74,51}
\definecolor{dimgray85}{RGB}{85,85,85}
\definecolor{gainsboro229}{RGB}{229,229,229}
\definecolor{lightgray204}{RGB}{204,204,204}

\begin{axis}[
axis background/.style={fill=gainsboro229},
axis line style={black},
height=3.5cm,
tick align=outside,
tick pos=left,
	title={Histogram of time for {\em in vivo}},
width=0.9\linewidth,
x grid style={lightgray204},
xlabel=\textcolor{dimgray85}{Time in seconds},
xmajorgrids,
xminorgrids,
xmin=0, xmax=63,
xtick style={color=dimgray85},
y grid style={lightgray204},
ylabel=\textcolor{dimgray85}{Number of clips},
ymajorgrids,
yminorgrids,
ymin=0, ymax=22.05,
ytick style={color=dimgray85}
]
\draw[draw=black,fill=chocolate2267451,very thin] (axis cs:0,0) rectangle (axis cs:3,21);
\draw[draw=black,fill=chocolate2267451,very thin] (axis cs:3,0) rectangle (axis cs:6,20);
\draw[draw=black,fill=chocolate2267451,very thin] (axis cs:6,0) rectangle (axis cs:9,13);
\draw[draw=black,fill=chocolate2267451,very thin] (axis cs:9,0) rectangle (axis cs:12,14);
\draw[draw=black,fill=chocolate2267451,very thin] (axis cs:12,0) rectangle (axis cs:15,7);
\draw[draw=black,fill=chocolate2267451,very thin] (axis cs:15,0) rectangle (axis cs:18,9);
\draw[draw=black,fill=chocolate2267451,very thin] (axis cs:18,0) rectangle (axis cs:21,8);
\draw[draw=black,fill=chocolate2267451,very thin] (axis cs:21,0) rectangle (axis cs:24,6);
\draw[draw=black,fill=chocolate2267451,very thin] (axis cs:24,0) rectangle (axis cs:27,2);
\draw[draw=black,fill=chocolate2267451,very thin] (axis cs:27,0) rectangle (axis cs:30,3);
\draw[draw=black,fill=chocolate2267451,very thin] (axis cs:30,0) rectangle (axis cs:33,4);
\draw[draw=black,fill=chocolate2267451,very thin] (axis cs:33,0) rectangle (axis cs:36,4);
\draw[draw=black,fill=chocolate2267451,very thin] (axis cs:36,0) rectangle (axis cs:39,4);
\draw[draw=black,fill=chocolate2267451,very thin] (axis cs:39,0) rectangle (axis cs:42,2);
\draw[draw=black,fill=chocolate2267451,very thin] (axis cs:42,0) rectangle (axis cs:45,0);
\draw[draw=black,fill=chocolate2267451,very thin] (axis cs:45,0) rectangle (axis cs:48,2);
\draw[draw=black,fill=chocolate2267451,very thin] (axis cs:48,0) rectangle (axis cs:51,1);
\draw[draw=black,fill=chocolate2267451,very thin] (axis cs:51,0) rectangle (axis cs:54,2);
\draw[draw=black,fill=chocolate2267451,very thin] (axis cs:54,0) rectangle (axis cs:57,0);
\draw[draw=black,fill=chocolate2267451,very thin] (axis cs:57,0) rectangle (axis cs:60,3);
\end{axis}

\end{tikzpicture}
 \caption{\Invivo{} experiment clip length distribution}
    \label{fig:histograminvivo}
\end{figure}
\textbf{\Invivo{}:}
For the \invivo{} experiments, we collect clips over 4 different porcine labs.
\Invivo{} actions are not specifically tuned for evaluating algorithm performance under specific movements.
Instead, they reflect actions needed to train surgeons.
This does not necessarily ensure diversity of actions, but more enables qualification of actions that are performed in a training scenario.
We do not prescribe any changes to the clinician workflow or procedure apart from adding instructions for switching to and from Firefly mode periodically.
Training scenarios include cholecystectomies (gall bladder removal) and a Nissen fundoplication (wrapping the top of the stomach around the esophagus).
The video-recorded procedures were carried out by expert personnel from Intuitive Surgical with hundreds of hours of experience testing surgical systems.
The length distribution of the clips recorded is shown in Fig.~\ref{fig:histograminvivo}.
These surgeon training operations can present different difficulties for algorithms
which include specularities, smoke from electrocautery tools, and blood.
For the tattooing process, a non-robotic laparoscopic instrument is used to pass the ink reservoir and the tattoo needle through an assist cannula
(a small 1 to 3cm port used for inserting materials such as gauze or sutures) into the surgical field, where they are picked up by robotic instruments. 
Tattooing is performed by using a robotic tool to hold each of the needle and reservoir separately and dipping the needle in the reservoir to coat it before tattooing each tissue point.
Visibility is confirmed by switching to Firefly (IR) mode.
The tattooing process takes five to ten minutes at the start of the procedure, and then the rest of surgical training is performed.
The procedure continues as normal with the surgeon being asked to switch between IR and visible light throughout their surgical training process to capture a diverse set of environments.
We additionally capture some specific scenes without surgical intervention that include periodic motion such as respiration or heart motion.

\begin{table}[tb]
\caption{Dataset statistics.}
\begin{center}
\begin{tabular}{c|ccc}
&Clips & Segmented tattooed points & Minutes\\
	\cmidrule{1-4}
\Invivo{} & 136 & 535 & 172.5\\
\Exvivo{} & 436 & 3069 & 42.1\\
\end{tabular}
\end{center}
\label{tab:datastats}
\end{table}

\subsection{Data Processing}
\label{sec:dataprocessing}

We separated each recorded continuous video sequence acquired as described above into many stand-alone samples to create our dataset,~\data{}.
Using the time-synchronized Firefly transition times recorded by the da Vinci Xi, we can capture the IR frames before and after the visible light mode by extracting frames with ffmpeg.
Fig.~\ref{fig:dataformatall} provides a useful visualization of the data format we now describe.
Given the unformatted videos along with calibration data, we convert the videos into clips that are bookended (having a frame on either side of start and finish) by single ground truth Firefly (IR) frames captured by the same camera system. 
For each clip there is a start IR image \(I_s \) and an end IR image \(I_e \), each having segmentations of the fluorescent ink \( S_s \) and \(S_e \), respectively, and the visible light video of said action \( V \).
All frames are of size 1280 \(\times\) 1024 pixels.
\(I_s, I_e\) are in Portable Network Graphic (png) format;
\(V\) is the action video in MPEG-4 Part 14 (mp4) format;
\(S_s, S_e\) are binary segmentations of the IR frames also in png format.
The binary images \( S_s ,  S_e \) are processed IR images containing  
\textit{segments} representing the ICG tattoos. 
These are STIR's ground truth labels and are obtained as described in the following paragraphs.

\textbf{Image Segmentation:}
To obtain the segments, we first take the videos and crop them according to the system's recorded transition times from visible light to IR and vice-versa.
This results in a visible light clip and two IR frames.
The IR images are then turned into binary segmentations to be used as ground truth by blurring the image, thresholding the IR channel, and then applying an opening transformation with a kernel size of 3.
This results in a smoothed set of \(n\) segments for each image.
These segments can vary in size between the start and end frame.
This could be amended by normalizing them all to be of fixed density using morphological filtering.
We opt not to normalize the size of segments since those that are closer to the camera, or that incidentally have more ink, should appear larger.
Instead, we filter using a fixed transformation to remove outliers or specularities.
Then, we use OpenCV to find segment centers and create bounding boxes around each segment.
The bounding boxes are used to assist the user annotation outlier filtering detailed in the next section.

\textbf{Segment Selection:}
Between a specific start and end frame, segments can appear and disappear due to motion or occlusion.
Additionally, specularities may appear as ICG segments in the binary images and have to be removed.
Therefore, we cannot include all segments from the start and end frames, because there would be segments that could be missed in either frame.
To account for this we observed the start and end binary frames for each clip, and we kept only segments that could be identified in both.
To identify these consistent segments, we selected the bounding boxes of all segments that are visible in both IR frames, by using VGG Image Annotator~\cite{duttaAnnotationSoftwareImages2019} and clicking inside each relevant bounding box.
With all co-visible segments labelled, we can remove the ones not present.
This results in the final segmentation images, \(S_s, S_e\), that are provided in our dataset.

In order to compute the 3D locations of each point, for each segment in the left image, we select the nearest point from the possible candidate segments in the right image on the same epipolar line (\(y \pm 5\mathrm{px}\)) using normalized cross correlation (NCC).
To verify correctness after each labelling session we ran a script that shows the paired start and end pairs and their filtered segmentations (example in Fig. \ref{fig:dataformatall}).
If the start and finish `constellations' looked different, then we repeated the image segmentation and selection steps outlined above or removed the sample from the dataset if unsuccessful.

\subsection{Dataset Details}
\label{sec:datadetails}

Table \ref{tab:datastats} summarizes the dataset statistics.
\data{} is provided as a set of numbered folders, with each folder, \textbf{\code{\textless \%03d\,\textgreater}},  containing:

\begin{itemize}
\item[] \code{left}
    \begin{itemize}
        \item[] \code{starticg.png} (ICG image of start frame)
        \item[] \code{endicg.png} (ICG image of end frame)
        \item[] \code{segmentation/startim.png}, 
        \item[] \code{segmentation/endim.png} (Filtered and segmented binary versions of ICG start and end image)
        \item[] \code{\textless ms\,\textgreater\_ms.mp4} (video file)
   \end{itemize}
\item[] \code{right}
    \begin{itemize}
        \item[] \code{starticg.png}
        \item[] \code{endim.png}
        \item[] \code{\textless ms\,\textgreater\_ms.mp4} (video file)
    \end{itemize}
\item[] \code{calib.json} Camera calibration parameters (intrinsics, relative stereo pose translation in metres and axis-angle rotation)
\end{itemize}

Tracking methods can use the left mp4 for purely-2D tracking, or they can use the stereo pair along with a depth estimation framework for 3D tracking evaluation.

\begin{figure}[tb]\centering
    \includegraphics[width=\linewidth]{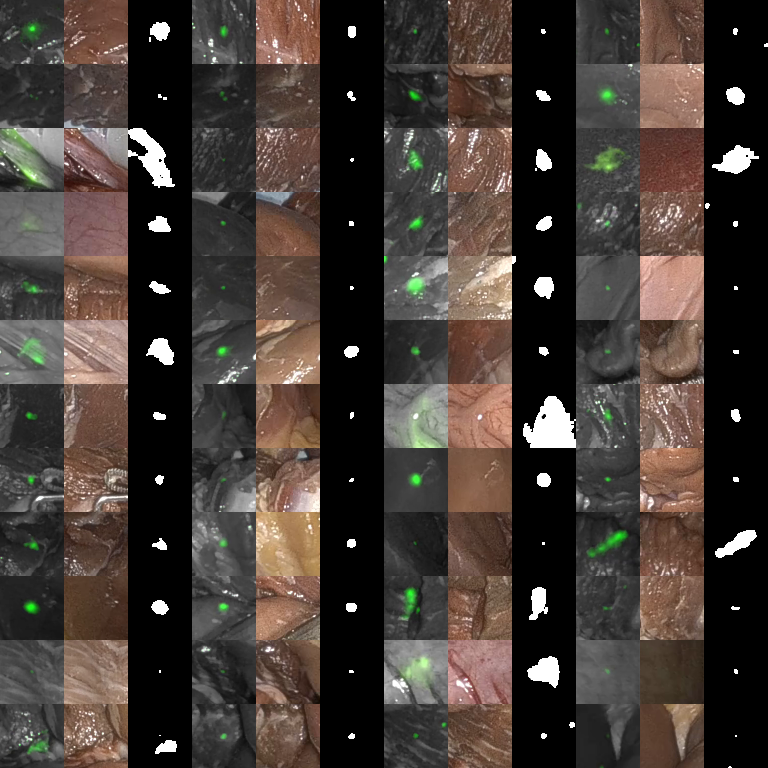}
	\caption{Four columns of ground truth triplets (48 triplets). IR image (left), visible light (middle), and segmentation (right) for each triplet. Note how the visible light images do not have visible ink that could interfere with algorithm performance.}
    \label{fig:patches}
\end{figure}

\textbf{Visibility of markers:}
We qualitatively verify that the marker segments are not visible under visible light by creating an automated sampling script to extract image patches.
After visually inspecting these segments we realized that in a small number of cases the segments are visible, but these cases are limited to the \invivo{} well-perfused organs.
Visibility of markers happens as the tattoo needle causes a small bleed in these organs that shows up as a spot on the tissue.
Hypothetically, we could use an image inpainting method to infill texture to cover the blood mark, but opt not to as this could introduce bias or other artifacts.
We do not believe the inpainting methods to be robust to bias since they introduce another pattern that could still be recognizable by a neural network, and we do not want to add any possible hallucinations into this environment.
In future work, the small bleeds could be resolved by using something with a clotting agent, or possibly a extra fine gauge needle.
See Fig.~\ref{fig:patches} for example randomly sampled image patches. 

\begin{figure}[tb]\centering
    \includegraphics[width=\linewidth]{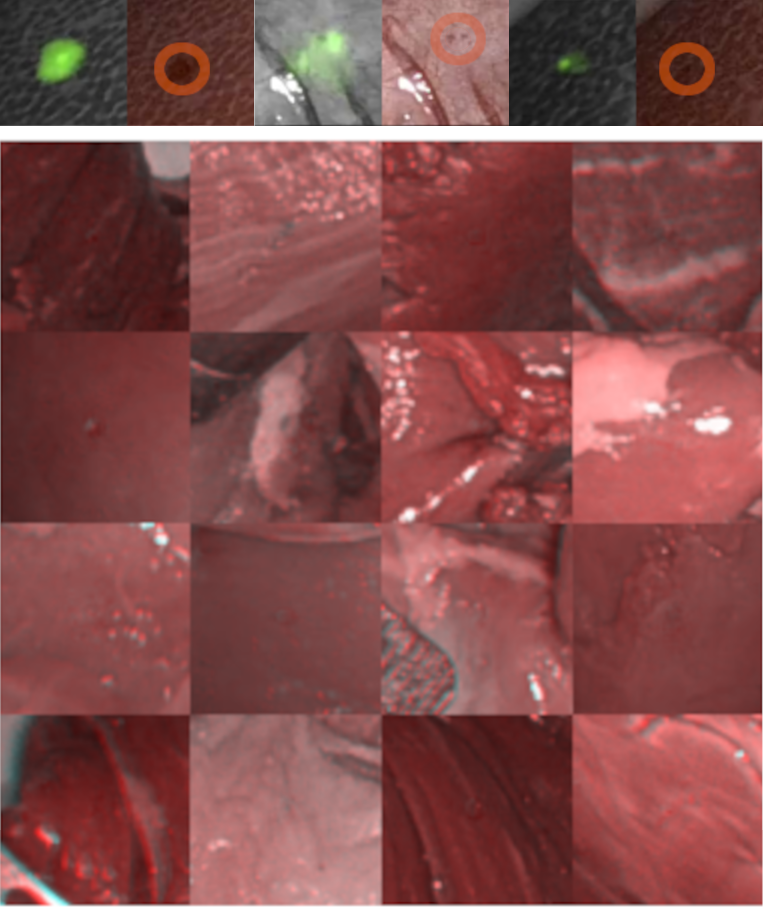}
	\caption{Top: 3 of 192 triplets that have visibility due to blood. Bottom: anaglyph of grayscale IR images overlaid on grayscale RGB images, demonstrating the extent of motion in IR transition. Cyan color offset from the red represents misalignment.}
    \label{fig:patchesandanaglyph}
\end{figure}

For details on our analysis, we performed two experiments to assess marker visibility.
(i) In a random set of 192 patches that is carefully inspected, 3 have noticeable visibility upon close inspection; these are also shown in Fig.~\ref{fig:patchesandanaglyph}.
Specifically, the blood is visible, and not the ICG itself.
The patches with visibility were only in the \invivo{} part of dataset.
We leave these points in the dataset, since we see it as important to quantify motion in samples on well-perfused organs such as the liver or kidneys, even if there might be some visibility present.
(ii) We additionally performed analysis on 292 random (\(21\times 21\))-sized samples to analyze texture features and see if we can cluster images according to this.
We compared Gray Level Co-occurence Matrix (GLCM) features~\cite{haralickTexturalFeaturesImage1973, ramolaStudyStatisticalMethods2020, zujovicStructuralTextureSimilarity2013} and saw no potential ways to cluster these patches into tattooed and non-tattooed regions.

\section{Evaluation and Results}
\label{sec:evaluationandres}
\subsection{Metrics for Evaluation}
\label{sec:evaluationmethods}
In order to quantify tracking results we use several metrics.
In this section we will explain the metrics used along with their benefits and drawbacks.
For each metric, the starting points are initialized with the start segmentation image as points (for endpoint error) or sets of points depending on the metric used.
The ending points are those in the end binary segmentation image, and the tracked (estimated) end points are tracked by tracking the start points frame-by-frame using each algorithm over the action video.
For a more in depth discussion of metrics, we recommend referring to Metrics Reloaded~\cite{maier-heinMetricsReloadedRecommendations2024}.
For more specific metrics used in tissue tracking, we recommend referring to SurgT~\cite{cartuchoSurgTChallengeBenchmark2023} and SENDD~\cite{schmidtSENDDSparseEfficient2023}.

\textbf{Endpoint Error}:
To calculate endpoint error, we calculate the smallest rectangular bounding boxes containing each segment.
The centers of bounding boxes are then treated as the segment center points.
Tracking performance is evaluated on these ground truth center points to the nearest corresponding point in tracking.
Endpoint error is the euclidean distance between the tracking evaluated on the video with initialization using the starting center points, and the ending center points.
Endpoint error can also be used to obtain a percentage metric of accuracy, by reporting the 
percentage of points with endpoint error under a certain threshold $\delta$.
The drawbacks of endpoint error lie in the fact that we are compressing a segment (set of pixels) into a single point sample.
Thus it would not be able to see if a region--which is small in our case--has scaled or warped incorrectly due to tracking.

\textbf{Chamfer Distance}:
The chamfer distance, \(d_{CD}\), between two sets of points is a measure of how close two regions are to one another by measuring the distance between each point in one set to the closest point in another set for each set.
Since the segmentation for each tattooed segment contains multiple pixels, we can use these pixels to evaluate chamfer distance.
Chamfer distance theoretically begins at \(0\) in the case of clips with zero motion, and increases with more motion error, or tracking drift.
Note that a pair of just two points that have a euclidean distance of 2 will have a chamfer distance of 4.
The primary issue with chamfer distance as a metric is that it is slow as it requires tracking all points in a segment.
The benefit is that for non-circular regions, chamfer distance does not smooth out the shape as endpoint error does.

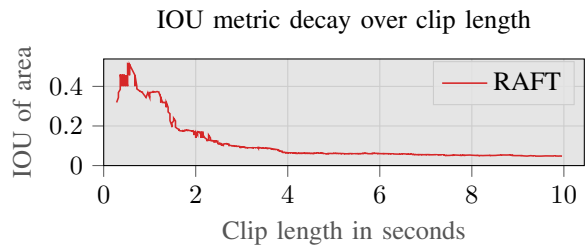
\begin{figure}[tb]
\centering
\begin{tikzpicture}

\definecolor{crimson2143940}{RGB}{214,39,40}
\definecolor{dimgray85}{RGB}{85,85,85}
\definecolor{gainsboro229}{RGB}{229,229,229}
\definecolor{lightgray204}{RGB}{204,204,204}
\definecolor{dimgray85}{RGB}{85,85,85}

\begin{axis}[
axis background/.style={fill=gainsboro229},
axis line style={black},
height=3cm,
tick align=outside,
tick pos=left,
title={IOU metric decay over clip length},
width=0.9\linewidth,
x grid style={lightgray204},
xmajorgrids,
xminorgrids,
xtick style={color=dimgray85},
y grid style={lightgray204},
ymajorgrids,
yminorgrids,
ytick style={color=dimgray85},
legend cell align={left},
legend style={fill opacity=0.8, draw opacity=1, text opacity=1, draw=lightgray204, fill=gainsboro229},
xlabel=\textcolor{dimgray85}{Clip length in seconds},
xmin=0, xmax=10.44395,
ylabel=\textcolor{dimgray85}{IOU of area},
ymin=0, ymax=0.538348935884484,
]
\addplot [semithick, crimson2143940]
table {0.281 0.318738229755179
0.32 0.343721193117198
0.32 0.368704156479218
0.36 0.384783112722367
0.36 0.400862068965517
0.36 0.459954844006568
0.4 0.400862068965517
0.4 0.459954844006568
0.44 0.400862068965517
0.44 0.459954844006568
0.48 0.400862068965517
0.48 0.455895515357076
0.52 0.400862068965517
0.52 0.455895515357076
0.52 0.510928961748634
0.52 0.514988290398126
0.56 0.510928961748634
0.56 0.455895515357076
0.56 0.510928961748634
0.68 0.455895515357076
0.68 0.400862068965517
0.68 0.455895515357076
0.72 0.400862068965517
0.72 0.387324761579116
0.8 0.373787454192714
0.8 0.371245805335966
0.8 0.373787454192714
0.84 0.371245805335966
0.84 0.368704156479218
0.92 0.343721193117198
0.96 0.368704156479218
1 0.343721193117198
1 0.368704156479218
1.08 0.371245805335966
1.08 0.373787454192714
1.12 0.371245805335966
1.16 0.373787454192714
1.2 0.371245805335966
1.2 0.368704156479218
1.24 0.343721193117198
1.24 0.318738229755179
1.24 0.343721193117198
1.28 0.318738229755179
1.28 0.318460023968499
1.32 0.318181818181818
1.32 0.318460023968499
1.36 0.318181818181818
1.36 0.29446208813021
1.36 0.318181818181818
1.4 0.29446208813021
1.4 0.270742358078603
1.44 0.261910710227223
1.44 0.253079062375844
1.44 0.227353274406729
1.48 0.201627486437613
1.48 0.194120031251261
1.48 0.201627486437613
1.48 0.213624450675785
1.52 0.225621414913958
1.52 0.239350238644901
1.56 0.225621414913958
1.56 0.213624450675785
1.56 0.201627486437613
1.56 0.194120031251261
1.6 0.186612576064909
1.64 0.182563791981901
1.64 0.178515007898894
1.64 0.17835853815788
1.64 0.178202068416866
1.64 0.176353931277281
1.64 0.174505794137696
1.64 0.176353931277281
1.72 0.174505794137696
1.72 0.173004875960669
1.72 0.171503957783641
1.72 0.173004875960669
1.72 0.174505794137696
1.72 0.176353931277281
1.76 0.174505794137696
1.76 0.173004875960669
1.76 0.174505794137696
1.76 0.176353931277281
1.76 0.178202068416866
1.76 0.17835853815788
1.8 0.178202068416866
1.8 0.176353931277281
1.8 0.178202068416866
1.84 0.176353931277281
1.84 0.174505794137696
1.84 0.176353931277281
1.84 0.178202068416866
1.88 0.176353931277281
1.92 0.174505794137696
1.92 0.173004875960669
1.92 0.171503957783641
1.92 0.173004875960669
1.96 0.171503957783641
1.96 0.17003291534667
1.96 0.168561872909699
1.96 0.154048378315315
1.96 0.168561872909699
2 0.154048378315315
2 0.13953488372093
2 0.154048378315315
2 0.168561872909699
2.04 0.154048378315315
2.04 0.168561872909699
2.04 0.17003291534667
2.04 0.171503957783641
2.08 0.17003291534667
2.08 0.168561872909699
2.12 0.154048378315315
2.12 0.13953488372093
2.12 0.154048378315315
2.12 0.168561872909699
2.12 0.17003291534667
2.16 0.168561872909699
2.16 0.154048378315315
2.16 0.13953488372093
2.16 0.154048378315315
2.2 0.13953488372093
2.24 0.133089810281518
2.24 0.126644736842105
2.28 0.125702406808769
2.28 0.124760076775432
2.28 0.125702406808769
2.28 0.126644736842105
2.28 0.133089810281518
2.28 0.13953488372093
2.28 0.154048378315315
2.32 0.13953488372093
2.32 0.133089810281518
2.32 0.126644736842105
2.36 0.125702406808769
2.36 0.124760076775432
2.36 0.123918499926177
2.36 0.123076923076923
2.36 0.123918499926177
2.36 0.124760076775432
2.36 0.125702406808769
2.4 0.124760076775432
2.4 0.123918499926177
2.4 0.123076923076923
2.4 0.116771019677996
2.4 0.123076923076923
2.4 0.123918499926177
2.44 0.123076923076923
2.44 0.116771019677996
2.44 0.123076923076923
2.44 0.123918499926177
2.48 0.123076923076923
2.48 0.116771019677996
2.48 0.11046511627907
2.48 0.108468466281497
2.48 0.11046511627907
2.52 0.108468466281497
2.52 0.11046511627907
2.52 0.113213458548812
2.56 0.11046511627907
2.56 0.108468466281497
2.56 0.106471816283925
2.56 0.103982176798679
2.56 0.101492537313433
2.56 0.103982176798679
2.6 0.101492537313433
2.6 0.103982176798679
2.6 0.106471816283925
2.6 0.108468466281497
2.64 0.11046511627907
2.68 0.108468466281497
2.68 0.11046511627907
2.72 0.108468466281497
2.72 0.106471816283925
2.72 0.103982176798679
2.72 0.101492537313433
2.76 0.100969482942431
2.76 0.100446428571429
2.76 0.100074720437602
2.76 0.099703012303776
2.76 0.100074720437602
2.76 0.100446428571429
2.8 0.100074720437602
2.8 0.099703012303776
2.84 0.0978372881424093
2.84 0.099703012303776
2.88 0.0978372881424093
2.88 0.0959715639810427
2.88 0.0951315516133552
2.88 0.0959715639810427
2.92 0.0978372881424093
2.96 0.0959715639810427
2.96 0.0951315516133552
2.96 0.0942915392456677
2.96 0.0939483646181999
2.96 0.0936051899907322
2.96 0.0918905715349555
3 0.0901759530791789
3 0.0897865212381341
3 0.0901759530791789
3 0.0918905715349555
3 0.0936051899907322
3 0.0939483646181999
3 0.0942915392456677
3.04 0.0939483646181999
3.04 0.0936051899907322
3.04 0.0918905715349555
3.04 0.0936051899907322
3.08 0.0918905715349555
3.08 0.0901759530791789
3.12 0.0897865212381341
3.12 0.0901759530791789
3.16 0.0897865212381341
3.16 0.0893970893970894
3.16 0.0897865212381341
3.2 0.0893970893970894
3.2 0.088958572361062
3.2 0.0893970893970894
3.24 0.088958572361062
3.24 0.0893970893970894
3.24 0.0897865212381341
3.24 0.0901759530791789
3.28 0.0897865212381341
3.28 0.0893970893970894
3.28 0.0897865212381341
3.32 0.0893970893970894
3.32 0.088958572361062
3.32 0.0885200553250346
3.32 0.088958572361062
3.359 0.0893970893970894
3.36 0.088958572361062
3.36 0.0893970893970894
3.36 0.0897865212381341
3.36 0.0901759530791789
3.36 0.0918905715349555
3.4 0.0901759530791789
3.4 0.0897865212381341
3.4 0.0893970893970894
3.44 0.088958572361062
3.44 0.0885200553250346
3.44 0.0881096859768681
3.48 0.0876993166287016
3.48 0.0881096859768681
3.52 0.0876993166287016
3.52 0.0858850565444393
3.52 0.0876993166287016
3.52 0.0881096859768681
3.56 0.0876993166287016
3.56 0.0858850565444393
3.6 0.0876993166287016
3.64 0.0858850565444393
3.64 0.084070796460177
3.64 0.0823614385561288
3.64 0.084070796460177
3.68 0.0823614385561288
3.68 0.0806520806520807
3.68 0.080267943013041
3.68 0.0798838053740015
3.68 0.080267943013041
3.68 0.0806520806520807
3.68 0.0813302244264587
3.68 0.0820083682008368
3.72 0.0813302244264587
3.72 0.0806520806520807
3.72 0.080267943013041
3.72 0.0798838053740015
3.72 0.080267943013041
3.72 0.0806520806520807
3.72 0.0813302244264587
3.76 0.0806520806520807
3.76 0.080267943013041
3.76 0.0798838053740015
3.76 0.080267943013041
3.8 0.0798838053740015
3.8 0.0784221754585546
3.8 0.0769605455431076
3.8 0.0751685844598655
3.8 0.0733766233766234
3.8 0.0729087841292566
3.8 0.0733766233766234
3.8 0.0751685844598655
3.84 0.0733766233766234
3.84 0.0729087841292566
3.84 0.0724409448818898
3.84 0.072302946667749
3.88 0.0721649484536082
3.88 0.0698724285647037
3.88 0.0675799086757991
3.88 0.0668735997810956
3.88 0.066167290886392
3.88 0.0661318569517497
3.88 0.066167290886392
3.92 0.0661318569517497
3.92 0.0660964230171073
3.92 0.0660297418515616
3.92 0.0659630606860158
3.92 0.0658968908445753
3.92 0.0659630606860158
3.92 0.0660297418515616
3.96 0.0659630606860158
3.96 0.0658968908445753
3.96 0.0658307210031348
3.96 0.0648302541185887
3.96 0.0638297872340425
4 0.0633126865755079
4 0.0638297872340425
4 0.0648302541185887
4.04 0.0638297872340425
4.04 0.0633126865755079
4.04 0.0627955859169732
4.04 0.0626477929584866
4.04 0.0627955859169732
4.04 0.0630809612753183
4.04 0.0633663366336634
4.08 0.0630809612753183
4.08 0.0627955859169732
4.08 0.0630809612753183
4.08 0.0633663366336634
4.08 0.063598061933853
4.12 0.0633663366336634
4.12 0.0630809612753183
4.12 0.0627955859169732
4.12 0.0626477929584866
4.16 0.0625
4.16 0.0618906685236769
4.16 0.0612813370473538
4.16 0.0604129977706121
4.16 0.0612813370473538
4.16 0.0618906685236769
4.2 0.0612813370473538
4.2 0.0618906685236769
4.2 0.0625
4.24 0.0618906685236769
4.24 0.0612813370473538
4.24 0.0618906685236769
4.28 0.0612813370473538
4.28 0.0604129977706121
4.28 0.0595446584938704
4.28 0.0604129977706121
4.28 0.0612813370473538
4.32 0.0618906685236769
4.32 0.0625
4.32 0.0626477929584866
4.36 0.0625
4.36 0.0618906685236769
4.36 0.0612813370473538
4.36 0.0604129977706121
4.36 0.0595446584938704
4.36 0.0604129977706121
4.36 0.0612813370473538
4.36 0.0618906685236769
4.4 0.0612813370473538
4.4 0.0604129977706121
4.4 0.0612813370473538
4.4 0.0618906685236769
4.44 0.0612813370473538
4.44 0.0604129977706121
4.44 0.0612813370473538
4.44 0.0618906685236769
4.48 0.0612813370473538
4.48 0.0604129977706121
4.48 0.0612813370473538
4.52 0.0604129977706121
4.52 0.0595446584938704
4.52 0.0604129977706121
4.52 0.0612813370473538
4.56 0.0604129977706121
4.56 0.0612813370473538
4.56 0.0618906685236769
4.56 0.0625
4.6 0.0626477929584866
4.6 0.0627955859169732
4.64 0.0626477929584866
4.64 0.0625
4.64 0.0626477929584866
4.68 0.0625
4.68 0.0626477929584866
4.68 0.0627955859169732
4.68 0.062996305969639
4.72 0.0627955859169732
4.72 0.0626477929584866
4.72 0.0625
4.72 0.0618906685236769
4.72 0.0625
4.76 0.0618906685236769
4.76 0.0612813370473538
4.76 0.0604129977706121
4.76 0.0612813370473538
4.8 0.0604129977706121
4.8 0.0595446584938704
4.8 0.0588195794405835
4.8 0.059445178335535
4.8 0.0594949184147027
4.84 0.059445178335535
4.84 0.0594949184147027
4.88 0.0595446584938704
4.88 0.0604129977706121
4.88 0.0612813370473538
4.92 0.0604129977706121
4.92 0.0595446584938704
4.92 0.0604129977706121
4.96 0.0612813370473538
5 0.0604129977706121
5 0.0595446584938704
5 0.0604129977706121
5 0.0612813370473538
5 0.0618906685236769
5.04 0.0612813370473538
5.04 0.0607158564936017
5.04 0.0612813370473538
5.08 0.0607158564936017
5.08 0.0601503759398496
5.08 0.0607158564936017
5.12 0.0601503759398496
5.12 0.05984751721686
5.12 0.0601503759398496
5.12 0.0607158564936017
5.16 0.0601503759398496
5.16 0.0607158564936017
5.2 0.0601503759398496
5.2 0.05984751721686
5.2 0.0595446584938704
5.2 0.05984751721686
5.2 0.0601503759398496
5.24 0.05984751721686
5.24 0.0595446584938704
5.24 0.0594949184147027
5.32 0.0595446584938704
5.32 0.05984751721686
5.32 0.0601503759398496
5.36 0.0607158564936017
5.4 0.0601503759398496
5.4 0.05984751721686
5.44 0.0595446584938704
5.44 0.0594949184147027
5.44 0.059445178335535
5.44 0.0594949184147027
5.44 0.0595446584938704
5.44 0.05984751721686
5.48 0.0601503759398496
5.52 0.05984751721686
5.52 0.0601503759398496
5.56 0.05984751721686
5.56 0.0601503759398496
5.64 0.0607158564936017
5.64 0.0612813370473538
5.64 0.0618906685236769
5.64 0.0625
5.68 0.0618906685236769
5.68 0.0612813370473538
5.68 0.0618906685236769
5.72 0.0612813370473538
5.72 0.0607158564936017
5.72 0.0601503759398496
5.72 0.05984751721686
5.72 0.0595446584938704
5.72 0.05984751721686
5.72 0.0601503759398496
5.84 0.05984751721686
5.84 0.0601503759398496
5.84 0.0607158564936017
5.84 0.0612813370473538
5.84 0.0618906685236769
5.84 0.0625
5.88 0.0618906685236769
5.88 0.0612813370473538
5.88 0.061044039986651
5.88 0.0612813370473538
5.92 0.0618906685236769
5.96 0.0612813370473538
5.96 0.061044039986651
5.96 0.0608067429259482
5.96 0.0604785594328989
5.96 0.0601503759398496
6 0.0604785594328989
6.04 0.0601503759398496
6.08 0.05984751721686
6.08 0.0601503759398496
6.08 0.0604785594328989
6.12 0.0601503759398496
6.2 0.05984751721686
6.2 0.0595446584938704
6.24 0.0594949184147027
6.24 0.059445178335535
6.28 0.0587698393614158
6.28 0.0580945003872967
6.28 0.0587698393614158
6.36 0.0580945003872967
6.44 0.0577308250728754
6.44 0.0580945003872967
6.48 0.0587698393614158
6.48 0.059445178335535
6.48 0.0594949184147027
6.52 0.059445178335535
6.56 0.0587698393614158
6.6 0.0580945003872967
6.6 0.0577308250728754
6.64 0.0573671497584541
6.64 0.0577308250728754
6.68 0.0573671497584541
6.68 0.0572550034506556
6.68 0.0571428571428571
6.68 0.0572550034506556
6.72 0.0571428571428571
6.72 0.0565217391304348
6.72 0.0559006211180124
6.76 0.055706671391697
6.76 0.0555127216653816
6.8 0.055072275322002
6.84 0.0546318289786223
6.84 0.055072275322002
6.84 0.0555127216653816
6.92 0.055072275322002
6.92 0.0546318289786223
6.96 0.055072275322002
7 0.0546318289786223
7 0.055072275322002
7.04 0.0555127216653816
7.08 0.055072275322002
7.08 0.0546318289786223
7.12 0.0544459593323605
7.16 0.0546318289786223
7.2 0.055072275322002
7.24 0.0546318289786223
7.24 0.0544459593323605
7.24 0.0546318289786223
7.28 0.0544459593323605
7.28 0.0542600896860987
7.319 0.0544459593323605
7.32 0.0542600896860987
7.32 0.0541570718700764
7.36 0.0540540540540541
7.36 0.0536693627934504
7.36 0.0540540540540541
7.399 0.0536693627934504
7.4 0.0532846715328467
7.4 0.0523245187519771
7.4 0.0513643659711075
7.4 0.0523245187519771
7.44 0.0513643659711075
7.44 0.0509402950799491
7.52 0.0513643659711075
7.56 0.0523245187519771
7.6 0.0513643659711075
7.6 0.0509402950799491
7.6 0.0513643659711075
7.6 0.0523245187519771
7.6 0.0532846715328467
7.6 0.0536693627934504
7.64 0.0532846715328467
7.72 0.0523245187519771
7.76 0.0513643659711075
7.76 0.0523245187519771
7.8 0.0513643659711075
7.8 0.0509402950799491
7.8 0.0513643659711075
7.8 0.0523245187519771
7.84 0.0513643659711075
7.84 0.0523245187519771
7.96 0.0532846715328467
8 0.0523245187519771
8.08 0.0513643659711075
8.12 0.0509402950799491
8.12 0.0505162241887906
8.12 0.050358513700821
8.12 0.0505162241887906
8.16 0.050358513700821
8.16 0.0505162241887906
8.2 0.0509402950799491
8.24 0.0505162241887906
8.28 0.0509402950799491
8.36 0.0513643659711075
8.4 0.0509402950799491
8.4 0.0513643659711075
8.44 0.0509402950799491
8.52 0.0513643659711075
8.56 0.0523245187519771
8.64 0.0532846715328467
8.64 0.0536693627934504
8.76 0.0532846715328467
8.8 0.0536693627934504
8.84 0.0532846715328467
8.84 0.0536693627934504
8.88 0.0532846715328467
8.88 0.0523245187519771
8.88 0.0513643659711075
8.88 0.0509402950799491
8.92 0.0505162241887906
8.96 0.050358513700821
9.04 0.0505162241887906
9.08 0.050358513700821
9.08 0.0502008032128514
9.12 0.0490662584678834
9.2 0.0479317137229153
9.24 0.0490662584678834
9.28 0.0502008032128514
9.32 0.0490662584678834
9.36 0.0479317137229153
9.36 0.0477753806709815
9.36 0.0479317137229153
9.48 0.0477753806709815
9.52 0.0479317137229153
9.56 0.0484103013059021
9.6 0.0488888888888889
9.8 0.0484103013059021
9.8 0.0479317137229153
9.8 0.0477753806709815
9.8 0.0479317137229153
9.84 0.0484103013059021
9.88 0.0479317137229153
9.92 0.0484103013059021
9.96 0.0479317137229153
};
\addlegendentry{RAFT}
\end{axis}

\end{tikzpicture}
 \caption{IOU decay on dataset. Other metrics are more robust in cases of larger motion relative to segmented region size.} \label{fig:ioudecay}
\end{figure}

\textbf{Intersection Over Union (IOU):}
We briefly discuss IOU, and after a preliminary analysis, omit its use for more in-depth quantification for reasons as follows.
IOU is defined as the intersection of two point sets (\(A,B\)) divided by the union of them.
\(A, B\) are the sets of finishing tracked points using the algorithm (every point that is a 1 in the binary image is tracked), and the end segments from the IR image, respectively.
We note that IOU is a poor metric for small segments like our tattooed markers.
Motions larger than a region's size frequently have an IOU of zero.
Like chamfer distance, this requires tracking and evaluating on full segments and thus can be slow.
Alternatively, segments can be approximated as larger rigid moving regions as in SurgT~\cite{cartuchoSurgTChallengeBenchmark2023}, but this assumes there is no local deformation.
See Fig.~\ref{fig:ioudecay} for a reference of IOU with respect to clip length.
Since IOU is noninformative in this case, we limit our evaluation of IOU to this figure.
Even short clips can have a very low IOU (\(<0.1\)).

\textbf{3D Error:}
Since we have regions from both the left and the right images, we can use these to get endpoint error or chamfer distance in 3D.
To obtain the 3D position of a segment center, we take segment centerpoints from the left segmented image, and find their depth.
We calculate depth by using NCC to match to the respective patch in the right image, and then verify this patch is an IR segment that lies in a valid depth range.
These matches are then backprojected into 3D.
They are tracked using a 3D tracking algorithm, or by tracking the point in each video, and projecting said point into space using a depth mapping algorithm.

\begin{figure*}[tb]
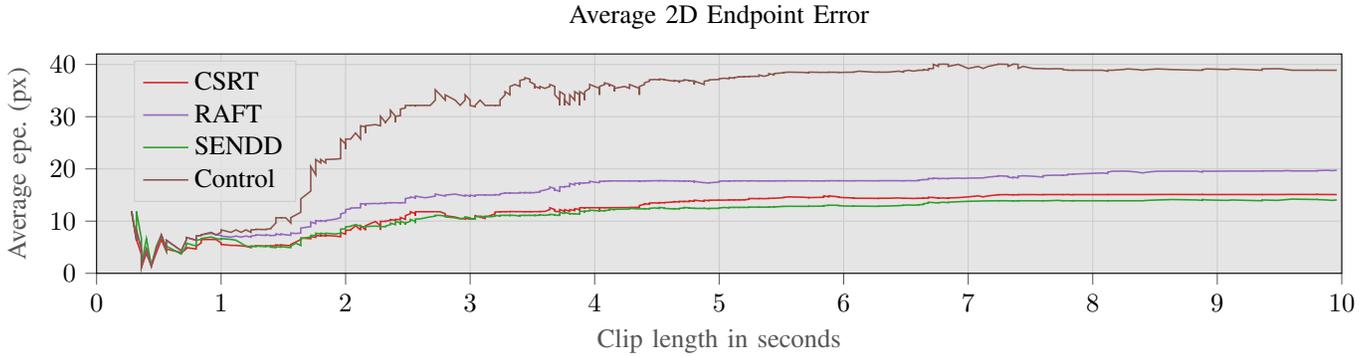

\centering

 \caption{Average 2D endpoint error on the full dataset. Error is the average for all clips up to length N.} \label{fig:2D_error}
\end{figure*}

\begin{figure}[tb]
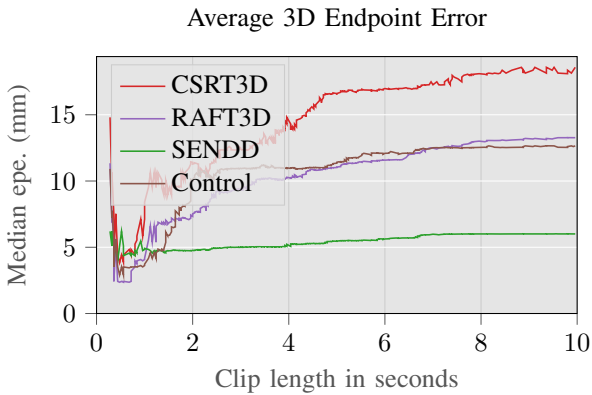

\centering
%
 \caption{3D endpoint error on the full dataset.} \label{fig:3D_error}
\end{figure}

\begin{figure}[tb]
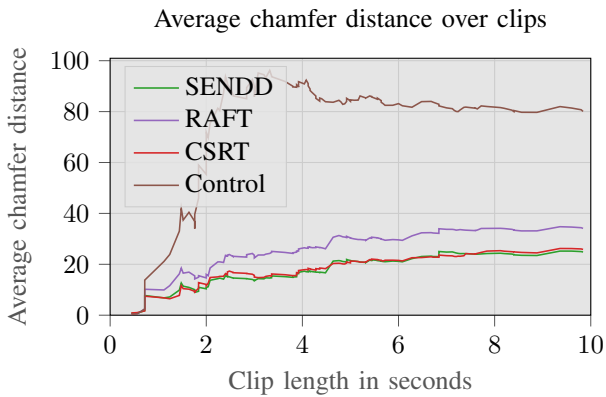

\centering
%
 \caption{Chamfer distance on a randomly sampled 128-clip subset. The curves show the average chamfer distance for all segments under the length on the x-axis.} \label{fig:chamfer_all}
\end{figure}

\subsection{Analysis of Trackers}
\label{sec:analysis}
We experiment with three different models (RAFT~\cite{teedRAFTRecurrentAllPairs2020}, CSRT~\cite{lukezicDiscriminativeCorrelationFilter2017}, and SENDD~\cite{schmidtSENDDSparseEfficient2023}), along with a control model which is a tracker that estimates no motion between each frame.
The control method effectively measures the distance between labelled start and end segments.
RAFT~\cite{teedRAFTRecurrentAllPairs2020} is an off-the-shelf optical flow model that uses a CNN along with all-pairs matching and a recursive iterative refinement scheme to estimate optical flow.
CSRT~\cite{lukezicDiscriminativeCorrelationFilter2017} is a classical correlation-based tracker for patches.
CSRT is chosen as a baseline due to its high performance in the SurgT challenge~\cite{cartuchoSurgTChallengeBenchmark2023}.
SENDD~\cite{schmidtSENDDSparseEfficient2023} is a graph-based tracking methodology trained in an unsupervised manner on surgical scenes (not in the STIR dataset) that uses sparse salient points to estimate motion anywhere in space.
We use these algorithms to showcase preliminary results on our dataset.
None of these models take into account occlusion or re-localization, so future methods with an underlying mapping or re-localization thread should outperform these in frequent cases of occlusion.
CSRT can be especially slow in some cases, as it scales according to how many regions to track, while SENDD is relatively independent, and RAFT is fixed in cost as it evaluates on every pixel anyways.
For evaluating short term frame-to-frame tracking methods we recommend estimating motion for each tracked point frame by frame pair over time.
If frame skips other than the default frame rate are used, they are noted as such.

\textbf{Whole-dataset model comparison:}
We first compare models on the whole dataset using endpoint error.
This is seen in Fig.~\ref{fig:2D_error} for 2D and Fig.~\ref{fig:3D_error} for 3D.
SENDD, RAFT, and CSRT can be seen in order of performance, with error increasing as clip length increases. 
For quantifying chamfer distance, we restrict experiments to a randomly sampled subset of 128 clips from the dataset as the CSRT evaluation takes >80x longer than the other methods as every pixel in a segmentation has to be tracked independently.
This takes over a week for CSRT to run.
Tracking results are shown in Fig.~\ref{fig:chamfer_all}.
This metric should give a closer sense of errors for irregularly shaped segments, with SENDD and CSRT being essentially identical in performance.

\begin{table}[tb]
	\caption{2D pixel endpoint error averages as percentage of image size by length. \(\pm\) denotes standard error of the mean. Normalized by image size (1280).}
\begin{center}
\begin{tabular}{c|ccc}
	\toprule
Model & Short & Medium & Long\\
	\cmidrule{1-4}
	CSRT & \textbf{1.40} $\pm$ 0.16\% & 2.11 $\pm$ 0.15\% & 3.34 $\pm$ 0.56\%\\
	RAFT & 1.73 $\pm$ 0.18\% & 2.36 $\pm$ 0.16\% & 3.86 $\pm$ 0.54\%\\
	SENDD & 1.43 $\pm$ 0.18\% & \textbf{1.93} $\pm$ 0.15\% & \textbf{2.81} $\pm$ 0.50\% \\
	Control & 3.26 $\pm$ 0.25\% & 3.99 $\pm$ 0.2\% & 4.22 $\pm$ 0.54\% \\
	\bottomrule
\end{tabular}
\end{center}
\label{tab:2Derrorbydiff}
\end{table}

\begin{table}[tb]
	\caption{3D millimetre endpoint error averages by length. \(\pm\) denotes standard error of the mean. Average working distance in surgery is 100 to 200mm.}
\begin{center}
\begin{tabular}{c|ccc}
	\toprule
Model & Short & Medium & Long\\
	\cmidrule{1-4}
	CSRT & 32.66 $\pm$ 6.59 & 73.34 $\pm$ 14.33 & 45.44 $\pm$ 6.54\\
	RAFT & 18.30 $\pm$ 3.58 & 36.30 $\pm$ 8.45 & 84.00 $\pm$ 38.06\\
	SENDD & \textbf{6.77} $\pm$ 0.50 & \textbf{8.81} $\pm$ 0.53 & \textbf{11.34} $\pm$ 1.59\\
	Control & 15.59 $\pm$ 2.59 & 16.33 $\pm$ 0.76 & 19.88 $\pm$ 2.27\\
	\bottomrule
\end{tabular}
\end{center}
\label{tab:3Derrorbydiff}
\end{table}

\textbf{Performance per length:}
We separate our dataset into three different sets according to their duration:
short (less than three seconds), medium (between three and seven seconds), and long (over seven seconds).
We note, long clips can still have easy cases, but overall the longer sequences should be more subject to drift and other errors.
For each length, we analyze algorithm performance and the standard error of the mean.
For the 2D error, see Table \ref{tab:2Derrorbydiff}.
The SENDD model outperforms the other methods in all cases except the short scenario, in which CSRT wins out.
For small motions, with little variation, CSRT likely has better smoothness constraints.
Performance for all methods clearly decays for the long clips compared to the short ones.
This emphasizes the importance of drift correction, occlusion management, and re-localization will hold in future works.

For the 3D error, see Table~\ref{tab:3Derrorbydiff}.
Since the CSRT and RAFT models track in separate frames without a proper sense of 3D, the error is higher in these cases.
This happens particularly for longer sequences as there is no consensus method between left and right tracks.
SENDD outperforms the other methods on all cases here.

\begin{table}[tb]
\caption{2D percentage of correctly tracked points with endpoint error < $\delta$ px.}
\begin{center}
\begin{tabular}{cc|cccc}
	\toprule
	Model & Length & 2 px & 5 px & 10 px & 15 px\\
	\cmidrule{1-6}
	SENDD & Short & 16\% & 46\%  & 70\%  & 79.9\% \\
	Control & Short		& 12\% & 27\% & 42\% & 49\% \\
	\cmidrule{1-6}
	SENDD & Medium & 6.7\% & 26\% & 53\% & 66\% \\
	Control & Medium	& 2.8\% & 10\% & 22\% & 33\% \\
	\cmidrule{1-6}
	SENDD  & Long  & 5.0\% & 24\% & 47\% & 62\% \\
	Control & Long		& 2.4\% & 13\% & 32\% & 44\% \\
	\bottomrule
\end{tabular}
\end{center}
\label{tab:2DPCK}
\end{table}

\begin{table}[tb]
\caption{3D percentage of correctly tracked points with endpoint error < $\delta$ mm.}
\begin{center}
\begin{tabular}{cc|cccc}
	\toprule
	Model & Length & 1 mm & 2 mm & 5 mm & 10 mm\\
	\cmidrule{1-6}
	SENDD	& Short		& 8.2\% & 25\% & 64\% & 84\% \\
	Control	& Short		& 8.7\% & 17\% & 37\% & 61\% \\
	\cmidrule{1-6}
	SENDD	& Medium	& 5.8\% & 19\% & 55\% & 78\% \\
	Control	& Medium	& 3.1\% & 9.5\% & 29\% & 51\% \\
	\cmidrule{1-6}
	SENDD	& Long		& 2.4\% & 14\% & 49\% & 72\% \\
	Control & Long		& 2.7\% & 10\% & 31\% & 59\% \\
	\bottomrule
\end{tabular}
\end{center}
\label{tab:3DPCK}
\end{table}

We perform an additional experiment to quantify performance of the best performer (SENDD) by evaluating accuracy at different thresholds~\cite{doerschTAPVidBenchmarkTracking2022} to investigate where it has difficulties.
For each difficulty level, Table~\ref{tab:2DPCK} shows the percentage of tracks that are within that threshold of ground truth.
We repeat the experiment with the 3D data, with results in Table~\ref{tab:3DPCK}.
A result of note is that the Control method has better performance for short (<3s) small distances (<1mm).
This could point to the importance of smoothness constraints and regularization in future work.
Since the Control method is given the ground truth 3D starting point (from both left and right images since it does not have a means of estimating depth), this can bias performance as it begins with the exact depth.
To explain this: for SENDD, we use the model-estimated depth for estimating the depth of the starting point.
This means for a action with no motion the control would have zero 3D error, but the SENDD model would have error relative to the quality of its depth estimation.

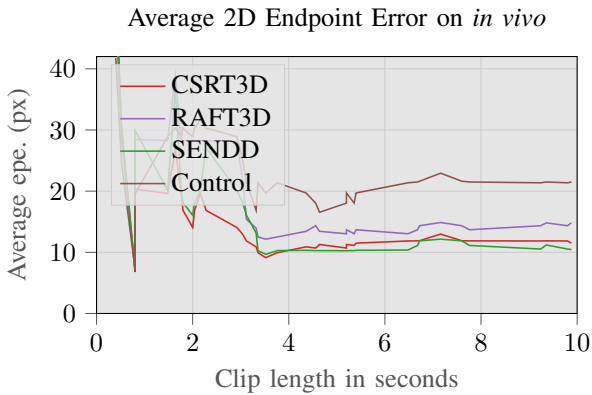
\begin{figure}[tb]
\centering
\begin{tikzpicture}

\definecolor{crimson2143940}{RGB}{214,39,40}
\definecolor{dimgray85}{RGB}{85,85,85}
\definecolor{forestgreen4416044}{RGB}{44,160,44}
\definecolor{gainsboro229}{RGB}{229,229,229}
\definecolor{lightgray204}{RGB}{204,204,204}
\definecolor{mediumpurple148103189}{RGB}{148,103,189}
\definecolor{sienna1408675}{RGB}{140,86,75}

\begin{axis}[
title={Average 2D Endpoint Error on \invivo{}},
axis background/.style={fill=gainsboro229},
axis line style={black},
height=5cm,
legend cell align={left},
legend style={
  fill opacity=0.8,
  draw opacity=1,
  text opacity=1,
  at={(0.03,0.97)},
  anchor=north west,
  draw=lightgray204,
  fill=gainsboro229
},
tick align=outside,
tick pos=left,
width=0.9\linewidth,
x grid style={lightgray204},
xlabel=\textcolor{dimgray85}{Clip length in seconds},
xmajorgrids,
xminorgrids,
xmin=0, xmax=10.,
xtick style={color=dimgray85},
y grid style={lightgray204},
ylabel=\textcolor{dimgray85}{Average epe. (px)},
ymajorgrids,
yminorgrids,
ymin=0, ymax=42.0097536714097,
ytick style={color=dimgray85}
]
\addplot [semithick, crimson2143940]
table {0.4 54.0832691319598
0.52 30.4261289500765
0.8 6.76898876819309
0.8 20.3163195548684
1.48 19.6355881826983
1.64 26.749619262121
1.76 19.6355881826983
1.8 16.8498317178274
2 14.0640752529565
2.04 16.8498317178274
2.16 19.6355881826983
2.28 16.8498317178274
2.92 14.0640752529565
3.04 12.9587111880758
3.12 11.853347123195
3.32 10.8934508617905
3.36 9.93355460038602
3.52 9.12439118451019
3.76 9.93355460038602
4.36 10.8934508617905
4.56 10.7008771254957
4.64 11.2771121243453
5.2 10.7008771254957
5.2 11.2771121243453
5.36 11.1455977951255
5.4 11.4994724591603
6.48 11.853347123195
6.68 11.8727792355141
6.72 11.8922113478332
7.16 12.9781433003949
7.6 11.8922113478332
7.76 11.8727792355141
9.24 11.853347123195
9.36 11.8727792355141
9.8 11.853347123195
9.88 11.4994724591603
};
\addlegendentry{CSRT3D}
\addplot [semithick, mediumpurple148103189]
table {0.4 49.4974746830583
0.52 28.4338803645365
0.8 7.37028604601469
0.8 28.4338803645365
1.48 28.2916256666975
1.64 38.8945501748779
1.76 28.2916256666975
1.8 28.2249532722138
2 28.15828087773
2.04 28.2249532722138
2.16 28.2916256666975
2.28 28.2249532722138
2.92 28.15828087773
3.04 21.7800780980865
3.12 15.4018753184429
3.32 13.955038432182
3.36 12.508201545921
3.52 12.1515293763186
3.76 12.508201545921
4.36 13.4269260701023
4.56 14.3456505942835
4.64 13.4269260701023
5.2 13.0327385728393
5.2 13.6891945835614
5.36 13.0327385728393
5.4 13.6891945835614
6.48 13.0327385728393
6.68 13.6891945835614
6.72 14.3456505942835
7.16 14.8737629563632
7.6 14.3456505942835
7.76 13.6891945835614
9.24 14.3456505942835
9.36 14.8111327564038
9.8 14.3456505942835
9.88 14.8111327564038
};
\addlegendentry{RAFT3D}
\addplot [semithick, forestgreen4416044]
table {0.4 53.6791908163525
0.52 30.4735755295744
0.8 7.2679602427964
0.8 29.9425261140749
1.48 20.160621798679
1.64 36.3888568920162
1.76 20.160621798679
1.8 18.1086623482179
2 16.0567028977568
2.04 18.1086623482179
2.16 20.160621798679
2.28 27.1328808970773
2.92 20.160621798679
3.04 18.1086623482179
3.12 16.0567028977568
3.32 13.1698282786352
3.36 10.2829536595136
3.52 9.660981724332
3.76 10.2829536595136
4.36 10.3307378720824
4.56 10.2829536595136
4.64 10.2752950620247
5.2 10.2676364645358
5.2 10.2752950620247
5.36 10.2829536595136
5.4 10.3307378720824
6.48 10.3785220846512
6.68 11.1243517845769
6.72 11.8701814845027
7.16 12.1541135237424
7.6 11.8701814845027
7.76 11.1243517845769
9.24 10.5329238045143
9.36 11.2015526445085
9.8 10.5329238045143
9.88 10.4557229445827
};
\addlegendentry{SENDD}
\addplot [semithick, sienna1408675]
table {0.4 41.7731971484108
0.52 24.5717415972128
0.8 7.37028604601469
0.8 19.5228419704761
1.48 28.9146937834482
1.64 30.2950458391929
1.76 28.9146937834482
1.8 30.2950458391929
2 28.9146937834482
2.04 30.2950458391929
2.16 31.6753978949376
2.28 30.2950458391929
2.92 28.9146937834482
3.04 25.1403164839796
3.12 21.365939184511
3.32 16.8824980951986
3.36 21.365939184511
3.52 19.702633309584
3.76 21.365939184511
4.36 19.702633309584
4.56 18.039327434657
4.64 16.5514173747761
5.2 18.039327434657
5.2 19.702633309584
5.36 18.039327434657
5.4 19.702633309584
6.48 21.365939184511
6.68 21.4954152924767
6.72 21.6248914004424
7.16 22.9375454578319
7.6 21.6248914004424
7.76 21.4954152924767
9.24 21.365939184511
9.36 21.4954152924767
9.8 21.365939184511
9.88 21.4954152924767
};
\addlegendentry{Control}
\end{axis}

\end{tikzpicture}
 	\caption{\textbf{2D} endpoint error on the \invivo{} dataset. Error is the average for all clips up to length N.} \label{fig:2D_invivo_error}
\end{figure}

\begin{figure}[tb]
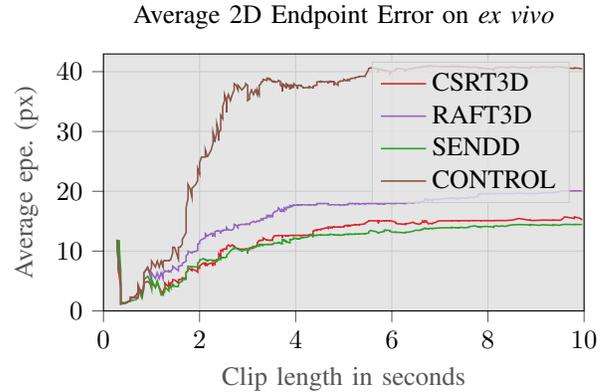

\centering

 \caption{\textbf{2D} endpoint error on the \exvivo{} dataset. Error is the average for all clips up to length N.} \label{fig:2D_exvivo_error}
\end{figure}

\textbf{\Invivo{} vs. \Exvivo}
We separate experiments by whether they are performed {\em in} or \exvivo{}.
For the \invivo{} experiments, the error is much more variant, likely due to more frequent occlusions and camera movement which these frame-level tracking methods do not account for.
Fig.~\ref{fig:2D_invivo_error} shows the 2D tracking endpoint error for \invivo{} experiments.
For the \exvivo{} experiments, SENDD also outperforms RAFT, CSRT and the Control methods (Fig.~\ref{fig:2D_exvivo_error}).

\begin{figure}[tb]
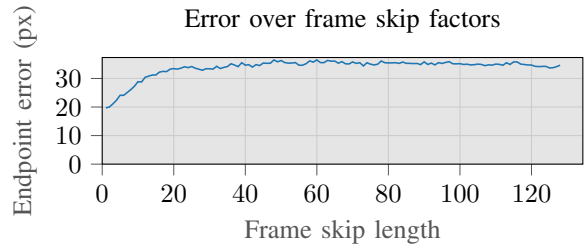

\centering
%
 \caption{Endpoint error over different frame skip factors when tracking on the whole dataset.} \label{fig:errorperskip}
\end{figure}

\textbf{Performance over different skip factors}:
We experiment with using different frame skip factors to see how this affects the tradeoff of efficiency and tracking performance on the dataset.
Skipping frames could reduce drift error, but could lead to failure in cases with large movement or fast changes in specularity.
Skipping frames could be used in tandem with a keyframe selection policy, or when occlusions are detected in the scene.
As seen in Fig.~\ref{fig:errorperskip}, performance decays until plateauing around twenty frames.

\section{Conclusion}
\label{sec:discussion}

With \data{}, we provide a novel means to quantify tracking algorithms.
Unlike other datasets, STIR uses physical markers which can reduce labelling bias, while also having the markers invisible, which can reduce algorithmic bias.
\data{} provides a modern dataset that  includes both \invivo{} and \exvivo{} samples.
We use multiple different frame to frame algorithms to evaluate performance on \data{} and show that among these methods, SENDD~\cite{schmidtSENDDSparseEfficient2023} has superior performance to the other frame-based tracking methods we tested.

\textbf{Limitations and future work}:
One limitation of \data{} is the temporal sparseness in sampling.
Since the ground truth segments are in IR and the camera takes 5 frames to transition, this means there could be small motions in the transition period leading to less accuracy in situations where there is large tissue motion that is not caused by the robot (since the robot remains fixed during the switch).
The measurement could be slightly off for heartbeat or respiration which can act as measurement noise.
Fig.~\ref{fig:patchesandanaglyph} demonstrates this.
Thus, in the future, movement could be quantified by having more frequent IR samples, or by quantifying in IR images.

Additionally, although \data{} is collected in multiple surgical scenarios, it does not contain non-laparoscopic interventions or human surgeries.
In the \invivo{} experiments, tissue types are limited to those present in abdominal interventions, so tissues such as muscle and lung are not included.
To enable non-porcine collection, we would likely have to reduce the additional time our labelling presents down from the 5 minutes or so it takes currently.
This would likely entail a streamlined process to apply ink to the needle rather than having to dip it using the robotic instruments.
Alternatively, we could design a specific instrument or a kinematic adjustment for this case since grasping a needle can be difficult.
Current difficulties include having to maintain grip on the needle or control the wrist motion of the grasper instrument normal to the needle, along with ensuring coverage when dipping the needle in the ink well.
Finally, using a finer gauge needle or a more viscous dye could help to reduce the small set of visible points on perfused organs even further.

Instrument masking and occlusion performance evaluation are other promising avenues.
Finally, evaluating on methods which provide re-localization or loop closure, or other long term point tracking is essential for long-term applications as it should help to further performance in situations where instruments occlude tissue.
In future work, recent deep learning-based techniques~\cite{harleyParticleVideoRevisited2022, neoralMFTLongTermTracking2024, wangTrackingEverythingEverywhere2023} that look extremely promising but have not been trained on surgical sequences will also be evaluated. 

\textbf{Other applications of \data{}}:
The point tattooing methodology in \data{} could be used for other applications which require tracking.
Tattooing in this precise form presented here, unlike submucosal injection of a bolus, is novel, and could be used for marking specific landmarks to return to.
For example, if we make an ultrasound measurement or biopsy at a specific point and want to mark that later in the video, we can tattoo and change to infrared every time we want to re-localize.
Additionally, if we have a preoperative registration, we could tattoo points and register them to the preoperative imaging, and use these tattooed points to maintain registration over time.
This differs from colorectal tattooing where a larger amount of dye is necessary to be able to localize outside the lumen (at the cost of accuracy).
Additionally, for marking points to suture, etc., this tattooing could be used as an alternative to cyanoacrylate beads~\cite{shademanFeasibilityNearinfraredMarkers2013}.

While not directly contemplated in the design of our experiments, the multitude of actions present in the STIR sequences may also enable a studies of surgical activity recognition~\cite{leaLearningConvolutionalAction2016, nwoyeCholecTriplet2021BenchmarkChallenge2023, vanamsterdamGestureRecognitionRobotic2022}.
The labelling of the data for this purpose is a time-consuming task that is left for future work

\textbf{Conclusion:}
We have presented a new methodology for accurately tracking tissue and used it to obtain a comprehensive dataset, \data{}, made public to enable better quantification of tracking methods.
Our dataset collection paradigm is novel and \data{} contains more sequences than the respective alternatives in surgical endoscopy.

\bibliography{datasetpaper}
\bibliographystyle{IEEEtran}

\end{document}